\documentclass[10pt,twocolumn,letterpaper]{article}

\usepackage[pagenumbers]{iccv} 
\usepackage{multirow}
\usepackage{amssymb} 
\usepackage{pifont}       
\usepackage{bbding}      
\usepackage{fontawesome}  
\usepackage{array} 
\usepackage[accsupp]{axessibility}

\newcommand{\best}[1]{\textcolor{red}{{\textbf{#1}}}}

\newcommand{\TopTwo}[1]{\textcolor{blue}{{\textbf{#1}}}}
\newcommand{\bestSmall}[1]{\textcolor{black}{{\textbf{#1}}}}

\definecolor{iccvblue}{rgb}{0.21,0.49,0.74}
\usepackage[pagebackref,breaklinks,colorlinks,allcolors=iccvblue]{hyperref}

\def\paperID{9853} 
\def\confName{ICCV}
\def\confYear{2025}

\title{MetaScope: Optics-Driven Neural Network for Ultra-Micro Metalens Endoscopy}

\author{Wuyang Li$^{1,3,}$\thanks{Equal Contribution. $^\dag$Corresponding Authors.} \qquad Wentao Pan$^{1,*}$\qquad
Xiaoyuan Liu$^{2,3,*}$ \qquad Zhendong Luo$^2$  \qquad Chenxin Li$^1$ \\ \qquad   Hengyu Liu$^1$ \qquad   Din Ping Tsai$^{2,\dag}$ \qquad  Mu Ku Chen$^{2,\dag}$ \qquad Yixuan Yuan$^{1,\dag}$ \\
$^1$The Chinese University of Hong Kong, $^2$City University of Hong Kong, $^3$EPFL\\
{Project Page: \href{https://cuhk-aim-group.github.io/MetaScope/}{https://cuhk-aim-group.github.io/MetaScope/}}
}
\begin{document}
\maketitle

\begin{abstract}
Miniaturized endoscopy has advanced accurate visual perception within the human body. Prevailing research remains limited to conventional cameras employing convex lenses, where the physical constraints with millimetre-scale thickness impose serious impediments on the micro-level clinical. Recently, with the emergence of meta-optics, ultra-micro imaging based on metalenses (micron-scale) has garnered great attention, serving as a promising solution. However, due to the physical difference of metalens, there is a large gap in data acquisition and algorithm research. In light of this, we aim to bridge this unexplored gap, advancing the novel metalens endoscopy. First, we establish datasets for metalens endoscopy and conduct preliminary optical simulation, identifying two derived optical issues that physically adhere to strong optical priors. Second, we propose MetaScope, a novel optics-driven neural network tailored for metalens endoscopy driven by physical optics. MetaScope comprises two novel designs: Optics-informed Intensity Adjustment (OIA), rectifying intensity decay by learning optical embeddings, and Optics-informed Chromatic Correction (OCC), mitigating chromatic aberration by learning spatial deformations informed by learned Point Spread Function (PSF) distributions. To enhance joint learning, we further deploy a gradient-guided distillation to transfer knowledge from the foundational model adaptively. Extensive experiments demonstrate that MetaScope not only outperforms state-of-the-art methods in both metalens segmentation and restoration but also achieves impressive generalized ability in real biomedical scenes.
\end{abstract}
    
\section{Introduction}
\label{sec:intro}

\begin{figure}[t]
\begin{center}
\includegraphics[width=1.0\linewidth]{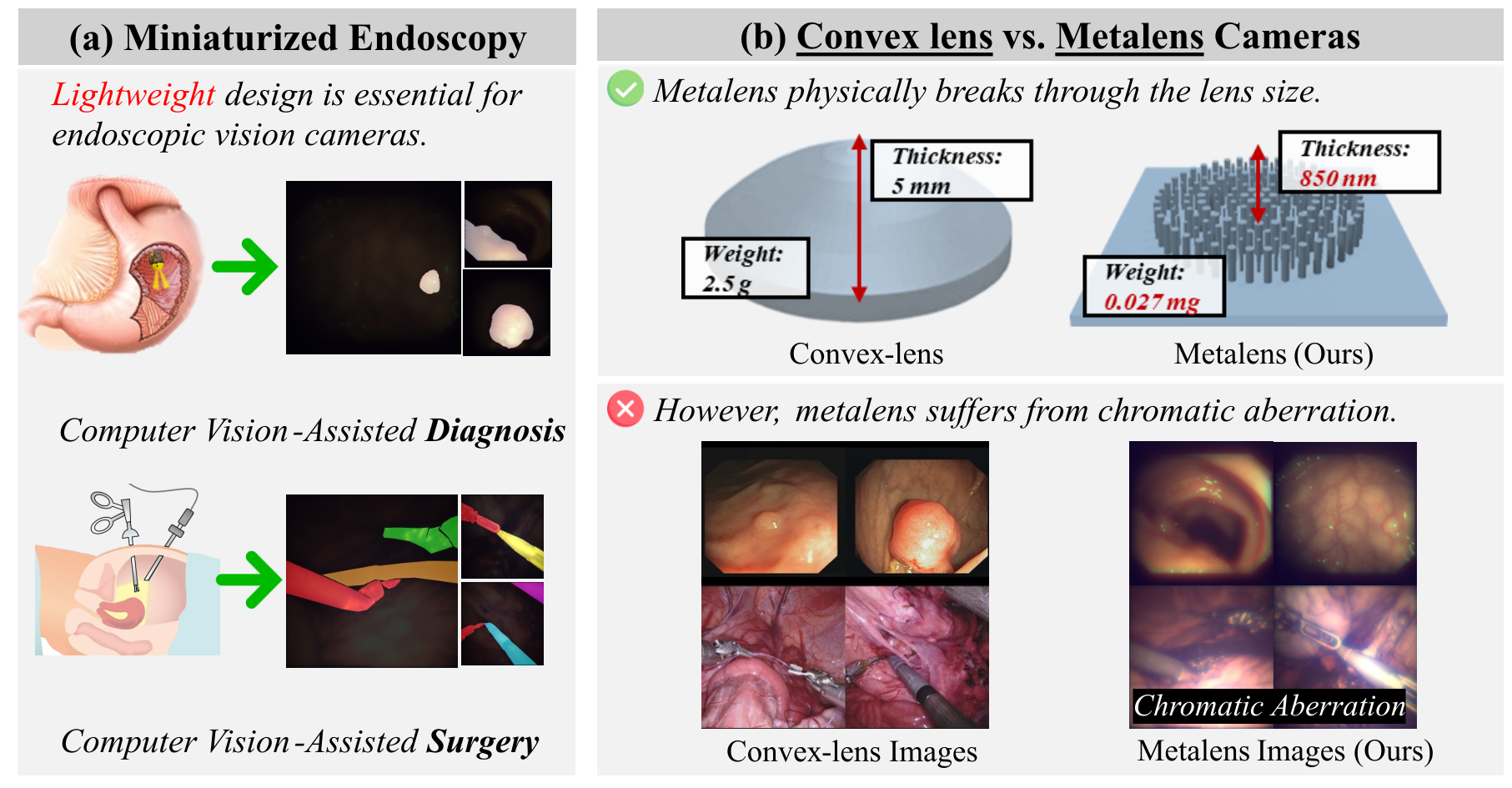}\vspace{-2pt}
\caption{(a) Miniaturized endoscopy; (b) Comparing convex-lens and our meta-lens about physical nature and imaging quality.}\label{fig:intro}
\end{center}\vspace{-20pt}
\end{figure}

Miniaturized endoscopy provides reliable internal visual perception within the human body (Fig.~\ref{fig:intro}a). Recent advancements in computer vision have significantly enhanced real-time clinical support, such as segmenting the gastrointestinal tract. This is vital for the early detection of serious diseases, aiding medical professionals in surgical decisions.

To facilitate endoscopic diagnosis and surgery~\cite{okagawa2022artificial,mori2019artificial,luo2025llm,liu2025comprehensive}, lots of efforts have been made to develop more robust, accurate, and efficient solutions for disease recognition and segmentation. Some works~\cite{jha2021kvasir,ali2019endoscopy,borgli2020hyperkvasir,li2024endora,li2024static,liu2025polyp} focus on data engineering by enhancing diversity, quality, and scale. Some other works~\cite{li2009computer,zhang2025ecsnn,fang2019selective,liu2024lgs,ali2024assessing,yang2023mrm} develop more advanced deep-learning models by addressing numerous algorithmic challenges, including domain generalization~\cite{ali2024assessing,li2022sigma,li2022scan}, privacy preservation~\cite{kaissis2020secure,chen2023medical,liu2022intervention}, and label-efficient learning~\cite{wu2023medical,li2023sigma++,liu2023decoupled}, among others.

Despite significant algorithmic advancements, existing studies are limited to conventional cameras that utilize convex lenses~\cite {lin2019achromatic,li2023revolutionary}, whose physical constraints pose substantial barriers to micro-level clinical applications. The inherent curvature limitations of convex lenses (Fig.~\ref{fig:intro}b left) hinder the development of ultra-lightweight profiles, restricting them to millimetre-scale thickness. This impairs navigation through narrow or curved anatomical structures and negatively impacts device size, power consumption, and operational efficiency, hindering progress toward micro-miniaturized clinical applications.

Recently, with the emergence of meta-optics~\cite{chen2022artificial,khorasaninejad2016metalenses,makarenko2022real,makarenko2024hardware,tseng2018metalenses}, ultra-micro imaging based on metalenses has garnered significant attention in micro-robotics, showing a promising solution~\cite{chia2024vivo,luo2021varifocal}. As shown in Fig.~\ref{fig:intro}b, these ultra-thin lenses utilize nanostructured arrays to manipulate light at subwavelength scales. In contrast to conventional in-vivo lenses, which typically weigh around \textbf{2.5 grams} with \textbf{5-millimetre} thickness, meta-lenses weigh less than \textbf{0.05 milligrams} and measure under \textbf{850-nanometer} in thickness. These superior physical properties significantly reduce the device size~\cite{liu2023underwater,liu2024meta,liu2024stereo}. This enables access to narrow anatomical structures and facilitates minimally invasive technologies within the human body.

While promising, these advancements pose challenges in maintaining imaging quality, as severe chromatic aberration~\cite{wang2018broadband,lin2019achromatic,chen2023meta} fundamentally limits their diagnostic reliability by distorting spectral signatures critical for tissue characterisation. The key challenge lies in achieving achromatic design, where the main difficulty stems from the wavelength-dependent phase modulation of nanostructures. Their resonant properties and material dispersion make it physically challenging to achieve consistent wavefront control across broad spectra, causing different wavelengths of light to focus at different points. Current physical solutions typically stack multiple layers~\cite{Li:20} or frequency-dependent phase engineering combining geometric-phase and propagation-phase elements~\cite{chen2018broadband}, but these face limitations in fabrication complexity, limited bandwidth, and trade-offs between efficiency and device size. Hence, \emph{we propose the computer vision solution for computationally achromatic meta-lens endoscopy (refer to Sec.~\ref{sec:dataset}),} underscoring its substantial industrial and academic value~\cite{tseng2018metalenses}.

\begin{figure*}[t]
\begin{center}
\includegraphics[width=0.85\linewidth]{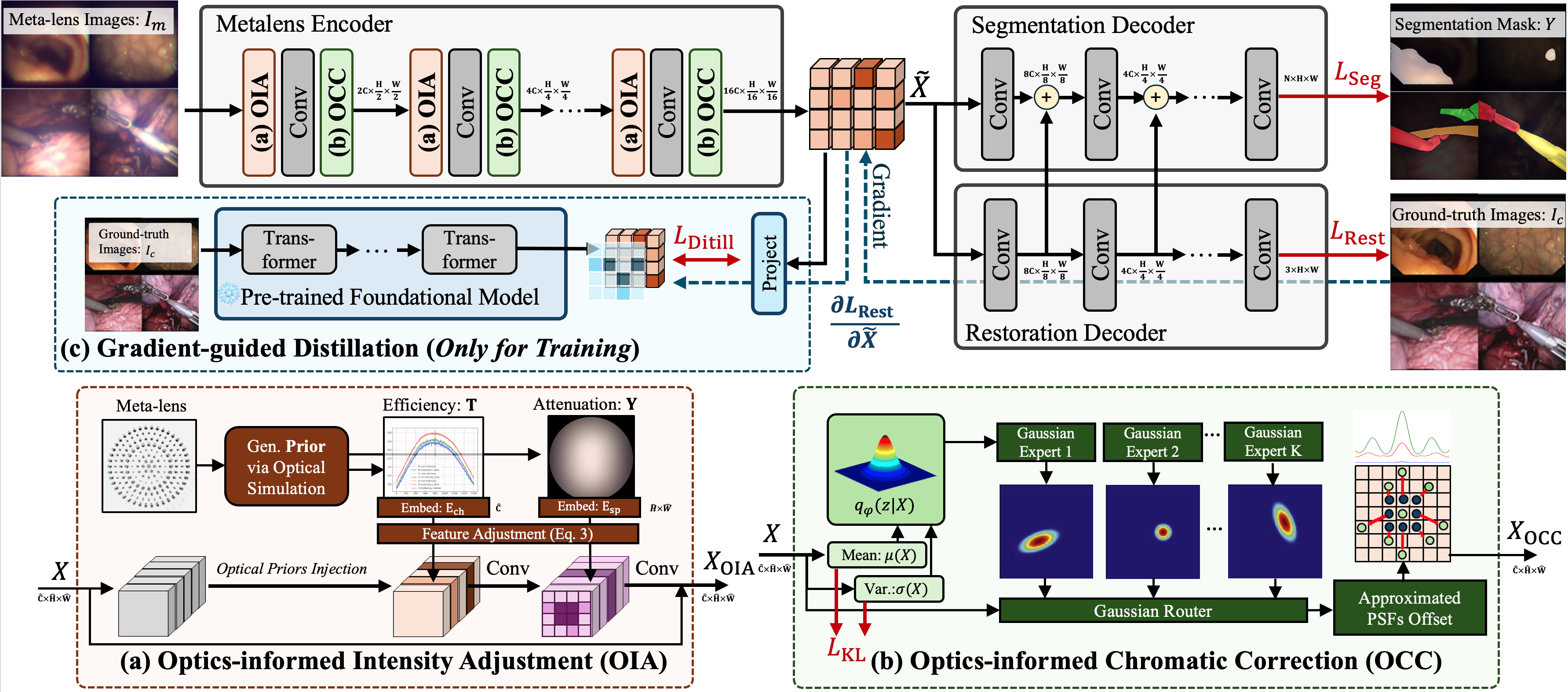}
\vspace{-5pt}
\caption{Illustration of the proposed MetaScope, which consists of two optics-informed modules, named (a) OIA and (b) OCC injected in the encoder to correct the meta-distortion, and a (c) gradient-guided distillation strategy to facilitate the restoration-aware representation. }\label{fig:overall}\vspace{-15pt}
\end{center}
\end{figure*}

To address this challenge, we delve into the nature of chromatic aberration with optical simulation, encapsulating it into two sub-issues (Sec.~\ref{sec:meta_dist}). \textbf{(1) Color and Position-Dependent Intensity Loss}: Different colors are attenuated unevenly with inherent optical prior. \textbf{(2) Color-Dependent Blur and Misalignment}: Different color has a varied ``blur kernel" (Point Spread Function, PSF). Mismatched color offsets create color fringing around objects, especially at high-contrast edges. Based on these observations, we further propose \textbf{MetaScope}, a novel optics-driven network to eliminate chromatic aberration and deliver precise segmentation for clinical applications. MetaScope employs a metalens encoder to correct distorted representations via Optics-informed Intensity Adjustment (OIA) and Optics-informed Chromatic Correction (OCC). Additionally, it incorporates a dual-branch decoder designed for accurate segmentation, enhanced by an auxiliary restoration branch (see Fig.~\ref{fig:overall}). Specifically, with optical simulations, OIA encodes optical priors into optical embeddings and then adjusts the attention map. OCC estimates PSF distributions by learning implicit distributions adaptively, eliminating chromatic aberrations through spatial aggregation. Furthermore, to enhance the robustness against chromatic aberration, we introduce gradient-informed distillation, which transfers foundational knowledge to enhance restored representation. \emph{By addressing both hardware and algorithmic barriers, we make clinical micro-miniaturization possible with enhanced physical efficiency.} The contributions are fivefold:

\begin{itemize}

\item We are the first to break through the endoscopy barrier via metalenses from both \textbf{data and algorithmic} aspects, illuminating ultra-micro applications of healthcare robotics. 

\item We propose five datasets for metalens-based diagnostics and surgery, starting and advancing the clinical research.

\item Delving into two imaging challenges with optical simulations, we propose a novel MetaScope to achieve joint restoration and segmentation guided by optical priors. 

\item The proposed MetaScope consists of an Optics-informed Intensity Adjustment to correct intensity attenuation with encoded optical embedding and an Optics-informed Chromatic Correction to rectify chromatic aberration. Besides, a gradient-informed distillation is proposed to enhance joint restoration-aware learning. 

\item Extensive experiments verify that MetaScope achieves SoTA segmentation and restoration, which also works impressively in real biomedical scenarios (Supp. C).

\end{itemize}

\section{Related Works}

\begin{figure*}[t]
\begin{center}
\includegraphics[width=1.0\linewidth]{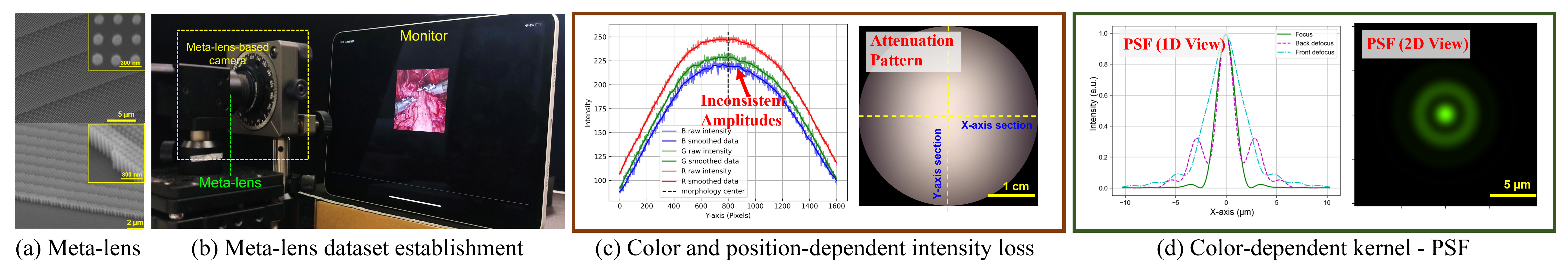}\vspace{-5pt}
\caption{(a) Our metalens nanostructure via scanning electron microscope; (b) Optical setup for meta-lens data collection; (c) Intensity loss in metalens imaging with color and spatial dependence; (d) Color-dependent blur and misalignment due to inconsistent PSF patterns.}\label{fig:two issues}
\end{center}
\vspace{-20pt}
\end{figure*}

\textbf{Metalens Imaging} uses optical metasurfaces to enhance the imaging capabilities of cameras~\cite{khorasaninejad2016metalenses,makarenko2022real,makarenko2024hardware,tseng2018metalenses}, which consist of arrays of nano-scale structures for precise focusing control, as shown in Fig.~\ref{fig:two issues}a. Their lightweight design and wide fields of view, \emph{make them highly suitable for applications in augmented reality, micro-robotics, and in-vivo healthcare}~\cite{chia2024vivo,luo2021varifocal}. However, manufacturing imperfections and inherent limitations lead to image distortion with severe chromatic aberration~\cite{ueno2024ai}. To break through this, artificial intelligence has been increasingly integrated into meta-lens imaging to facilitate meta-lens design~\cite{an2022deep, zhelyeznyakov2021deep, phan2019high, jiang2019free, wen2020robust} and enhance image quality through deep models~\cite{dong2024achromatic, tseng2021neural, hsu2023high, seo2023deep}. Unlike existing studies focusing on restoration, we are the first to explore micro-miniaturized endoscopic applications, proposing a unified perception model driven by optical priors.

\noindent\textbf{Medical Image Segmentation}~\cite{azad2024medical,liu2024diffrect,li2024flaws} autonomously delineates intricate anatomical structures, streamlining clinical procedures. With the U-Net architecture~\cite{ronneberger2015u}, numerous variants~\cite{isensee2021nnu,cciccek20163d,wang2019dense,chen2021transunet,wuyang2021joint,pan2023semi,hering2022learn2reg,pan2023human,lin2025rethinking,yan2021hierarchical} have been developed to enhance multi-scale fusion, skip connections, and bottleneck features~\cite{isensee2021nnu}. Recently, U-shaped models such as UNeXt~\cite{valanarasu2022unext}, DC-UNet~\cite{lou2021dc}, Rolling-Unet~\cite{liu2024rolling}, U-Mamba~\cite{ma2024u}, and U-KAN~\cite{li2024u} have improved model efficiency and interoperability. Additionally, generic segmentation methods have shown remarkable efficacy in medical applications, such as Mask2Former~\cite{cheng2021mask2former}, EDAformer~\cite{yu2025embedding}, and SegNext~\cite{guo2022segnext}. Unlike traditional methods that rely on high-fidelity images, our approach targets the more challenging metalens images via optical observations.

\noindent\textbf{Image Restoration}~\cite{kim2016accurate,dai2019second,chen2022simple,liang2021swinir,lin2023diffbir} seeks to reconstruct high-quality images from degraded inputs, such as rainy, foggy, etc~\cite{chen2024bidirectional,yu2025embedding}. Chen \emph{et al.}~\cite{chen2022simple} proposes a comprehensive study of restoration-oriented CNN design and introduces NAFNet, a lightweight model for efficient image restoration. With the scalability of the Transformers architecture, SwinIR~\cite{liang2021swinir} employs shifted-window attention to enhance global context while preserving local details. Additionally, some methods, such as Restormer~\cite{zamir2022restormer}, SFHformer~\cite{jiangfast}, and MambaIR~\cite{guo2025mambair} have gained attention due to their novel network architectures. Instead of focusing on the natural restoration with convex lenses, we investigate advanced meta-lens imaging driven by distinct optical simulation and explore its potential to assist perception tasks.

\section{Preliminaries}
\label{sec:preliminaries}

\textbf{Metalens Imaging.} Conventional lenses achieve light focusing by varying thickness in a convex geometry and using multiple lens elements, which poses challenges for micron-miniaturized practice due to the bulky optical systems. In contrast, \textbf{metalenses} employ nanoscale structures, meta-atoms patterned on a flat surface (Fig.~\ref{fig:two issues}a) to manipulate the light wavefront in a much more compact way. Each meta-atom provides specific amplitude and phase modulation. The focusing phase profile can be expressed as:
\begin{equation}\label{eq:meta_imaging}
\phi(r, \lambda) = -\left[\frac{2\pi}{\lambda} \left(\sqrt{r^2 + f^2} - f\right)\right], 
\end{equation}
where $\phi(r, \lambda)$ is the required phase for the meta-atom at position $r$ for wavelength $\lambda$, and $f$ is the desired focal length. Without achromatic design, metalenses exhibit wavelength-dependent focal lengths, expressed as $f(\lambda)$. \emph{This variation prevents different wavelengths from converging on a single imaging plane.} We designed and fabricated polarization-independent gallium nitride (GaN) metalens with a diameter of 2.6 mm and a focal length of 10 mm.

\subsection{Metalens Dataset Establishment}\label{sec:dataset}
Due to the limited imaging quality of metalens cameras, a significant barrier exists to clinical practice, resulting in insufficient endoscopic data collection. This scarcity of data impedes the design of algorithms aimed at correcting chromatic aberration. To overcome this, we employ a proxy photographic solution to establish datasets, as shown in Fig.~\ref{fig:two issues}b. \emph{As the chromatic aberration patterns adhere to consistent physical principles, we can create paired metalens mapping data with and without distortion to train the model that approximates the restoration function.} Based on this insight, we construct the necessary pairs by projecting images from publicly available endoscopic datasets onto a screen and capturing images with a metalens camera. Thus, we can generate the metalens dataset with a triple data format, including the metalens image, associated high-quality ground truth, and the corresponding semantic labels, such as segmentation. Note that we empirically find that the model trained on our datasets exhibits \textbf{generalization capacity to real scenes with varied depth and different metalenses}, demonstrating significant value to the community. More details can be found in the Supp C.

\subsection{Identifying Metalens Imaging Challenges}
\label{sec:meta_dist}
The challenge of metalens images arises from the wavelength dependence of the focal length, which fails to focus all light onto a single focal plane. To explore this unique characteristic, we conduct optical simulations and identify two derived issues, motivating our method designs.

\noindent\textbf{(1) Color and Position-Dependent Intensity Loss.} 
We illuminate the metalens with the directional white light of uniform intensity while the CMOS sensor applies gains of 1.67, 1, and 2.34 to RGB channels (built-in white balance function). We analyze the resulting intensity variations in the captured images, as shown in Fig.~\ref{fig:two issues}c, summarizing two observations: (a) Imbalanced RGB Channel Intensities: Even under the automatic white balance, color channels exhibit inconsistent brightness due to wavelength-dependent responses of meta-atoms. Specifically, non-optimized wavelengths (e.g., red/blue) suffer from simultaneous amplitude and phase mismatch in meta-atoms, significantly reducing modulation efficiency and causing light intensity loss. (b) Vignetting and Structured Noise: The surrounding regions experience significantly greater attenuation compared to the central area, which is analogous to the \emph{dark channel} observed in de-hazing~\cite{he2010single}. This is because of the superposition of off-axis efficiency drop of nanostructures $\eta_{\text{meta}}(\theta)$ and traditional geometric vignetting (\(\cos^4\theta\) law), which could be approximated as $I_{\text{total}}(\theta) = \eta_{\text{meta}}(\theta) \cdot \cos^4\theta \cdot I_0$. Inhomogeneous amplitude responses of different meta-atoms and nanofabrication defects induce local random fluctuations in transmitted light intensity, manifesting as spatially correlated brightness noise. \underline{\emph{\textbf{Motivation:}}} To fully leverage observed optical priors, we propose to encode the efficiency and radial priors to adjust the feature learning with restoration awareness (Sec.~\ref{sec:method_part1}).

\noindent\textbf{(2) Color-Dependent Blur and Misalignment.} PSF is the imaging response of an optical system to an ideal point light source, directly governing spatial resolution and color fidelity, which could be approximated as $I_{\text{output}} = \sum_{\lambda} \left[ T(\lambda) \cdot \left( \eta_{\text{meta}}(\theta) \cdot \cos^4\theta \cdot I_0(\lambda) \ast \text{PSF}(\lambda, x, y) \right) \right] + \epsilon$, where $ \epsilon$ is noise. Fig.~\ref{fig:two issues}d shows the simulated PSFs of a coherent plane wave (ideal light source) under the wavelength of 532 nm (green). In metalenses, chromatic aberrations cause wavelength-dependent PSF variations. Each wavelength (color) has a different focal length for a sensor fixed at the design wavelength (e.g., green), shorter (red) and longer (blue) wavelengths defocus, broadening their PSFs. Blue/red edges appear more blurred than green ones, akin to applying color-specific Gaussian kernels. Imperfect phase gradient compensation of the metalens results in PSF centroid displacement across wavelengths. Hence, high-contrast edges exhibit color fringing, as misaligned RGB PSFs create artificial chromatic gradients. Our observations indicate that such blur and fringing cause the signal to spread from the source pixel into adjacent pixels. \underline{\emph{\textbf{Motivation}:}} To tackle this issue, we propose a novel solution by estimating the full-color PSF \textbf{distribution} via learning inherent Gaussian mixtures adaptively, thereby spatially guiding correction of chromatic aberrations (Sec.~\ref{sec:method_part2}).

\noindent\textbf{Problem Formulation.} We have triplet data $\{I_m, {I}_c, {Y}\}$ for training, including degraded images captured by metalens $I_m$, the associated high-quality ones captured from the convex lens ${I}_c$, and the segmentation mask ${Y}$. Our goal is to train a unified model $\mathcal{F}$ that can achieve reliable perception $\Tilde{Y}$ and restoration ${I}_c$ using metalens images $I_m$. In the test stage, the model can predict restored images and masks directly from metalens images: $\{\Tilde{Y},\Tilde{I}_c\} = \mathcal{F}(I_m)$.

\section{Methodology}
\label{sec:method}

As shown in Fig.~\ref{fig:overall}, given metalens images $I_m$, we employ a metalens encoder $\mathcal{F}_{\text{enc}}$, which consists of two key modules, (a) OIA and (b) OCC, to generate the restoration-aware latent $\Tilde{X}$, addressing the two issues in chromatic aberration (Sec.~\ref{sec:meta_dist}) with optical priors. Then, $\Tilde{X}$ is sent to a dual-branch decoder $\mathcal{F}_{\text{dec}}$ to generate both the restored image $\Tilde{I}_m$ and the segmentation mask $\Tilde{Y}$. In this process, the intermediate feature of meta-restoration (low-level) guides the segmentation branch with a simple residual connection. During the training, we implement (c) gradient-guided distillation that encodes the convex-lens image $I_c$ with the foundation model to facilitate the learning of restoration.

\subsection{Optics-informed Intensity Adjustment}
\label{sec:method_part1}

Metalens images experience severe attenuation in adhering to specific priors. However, the limited data scale makes it challenging to learn these priors solely through data fitting. To overcome this, we aim to adjust the feature using the optical priors derived from simulations, as shown in Fig.~\ref{fig:overall}a.

\noindent\textbf{Optical Prior Encoding.} We begin by sending the metalens image $I_m$ through a metalens encoding block to obtain the feature ${\widetilde{X}}$. Due to inconsistent wavelengths, the captured images $I_m$ exhibit varying intensity attenuation with optical evidence. To tackle this, we use the optical simulation to generate two types of optical priors (see Sec.~\ref{sec:meta_dist}): the color-depended vector $\mathbf{T}\in \mathbb{R}^{L}$ encoding the efficiency of $L$ visible light in sequential wavelengths, and spatial attenuation prior $\mathbf{Y}\in \mathbb{R}^{H\times W}$. Note that those priors can be generalized to different metalens due to the similar attenuation patterns. 
Then, we encode them into the optical embedding at the channel $\mathbf{E}_{\text{ch}}\in \mathbb{R}^{\hat{C}}$ and spatial $\mathbf{E}_{\text{sp}}\in \mathbb{R}^{\hat{H} \times \hat{W}}$ levels to calibrate the input feature ${X}\in \mathbb{R}^{\hat{C}\times \hat{H} \times \hat{W}}$ :
\begin{equation}\label{eq:optical embedding}
\mathbf{E}_{\text{ch}} = \mathcal{F}^{\text{Enc}}_{\text{Fc}} \left( \mathbf{T} \right);
\mathbf{E}_{\text{sp}} = \mathcal{F}^{\text{Enc}}_{\text{Conv}} \left( \mathbf{Y} \oplus \mathbf{C}_{x} \oplus \mathbf{C}_{y} \right),
\end{equation}
where $\mathbf{C}_{x/y}\in \mathbb{R}^{H \times W}$ indicate the normalized coordinates, $\oplus$ is the concatenation, and $\mathcal{F}^{\text{Enc}}_{\text{Fc}}/\mathcal{F}^{\text{Enc}}_{\text{Conv}}$ are simple fully connected and convolutional based layers, abbreviated as $\texttt{Conv}/\texttt{Fc}$ in following sections. These optical embeddings provide significant optical guidance to correct the intensity attenuation at the spatial and channel levels.

\noindent\textbf{Feature Adjustment with Optical Prior.} To integrate optical prior, a natural idea is to inject embedding into the network like diffusion models~\cite{ho2020denoising}. However, instead of sharing across all features, metalens images introduce attenuation patterns exhibiting variability across wavelengths and positions. Hence, we consider the distinct physical properties of efficiency (color-based) and spatial priors (position-based) and propose a simple yet effective method that uses optical embedding to rectify channel and spatial attention. Specifically, the adjusted attention is integrated as follows:
\begin{equation}\label{eq:feature_adjustment}
\begin{aligned}
\text{Attn}_{\text{ch}}{(X)} &= \text{Sigmoid}\left( \mathcal{F}^{\text{Proj}}_{\text{Fc}}\left( \text{Mean}_{\text{sp}}(X) + \mathbf{E}_{\text{ch}} \right) \right), \\
\text{Attn}_{\text{sp}}{(X)} &= \text{Sigmoid}\left( \mathcal{F}^{\text{Proj}}_{\text{Conv}}\left( \text{Mean}_{\text{ch}}(X) + \mathbf{E}_{\text{sp}} \right) \right),
\end{aligned}
\end{equation}
where $\text{Mean}_{\text{sp/ch}}$ indicates the averaging over the spatial and channel dimension, and $ \mathcal{F}_{\text{Fc/Conv}}^{\text{Proj}}$ is a projection layer. Then, given the feature $X$, the adjusted one can be represented as: $X_{\text{OIA}} = \texttt{Conv}(\text{Attn}_{\text{sp}}{(X)} \odot(\texttt{Conv}(\text{Attn}_{\text{ch}}{(X)}\odot X))) + X$, where $\odot$ is Hadamard product. Thus, by embedding the optical prior, the model can dynamically adjust representations based on the color-specific and spatially varying optical prior, improving the restoration-aware encoding.

\subsection{Optics-informed Chromatic Correction}
\label{sec:method_part2}

Metalens images suffer from chromatic artifacts due to wavelength-specific PSFs. Considering the difficulty in estimating explicit PSFs. we model the implicit PFS distribution via Gaussian Mixture Models (GMM) in an adaptive manner for chromatic correction (Fig.~\ref{fig:overall}b).

\noindent\textbf{Formulating Implicit PSF Distribution.} Given an input light with wavelength $\lambda$, metalens projects it to the 2-D image at a specific position $\mathbf{p}=[x,y]$ coupled with dispersed components to the neighbouring pixels, which can be formulated as the spatial dispersion $\Delta \mathbf{p}=[\Delta x, \Delta y]$, adhering the implicit PSF distribution $P(\Delta{\mathbf{p}})$. As shown in Fig.~\ref{fig:two issues} (d), such dispersion can be naturally modeled as a mixture of Gaussians with a main lobe accompanied by symmetric side lobes. Based on this observation, we formulate the implicit PSF distribution with GMM as follows,
\begin{equation}\label{eqn_gmm}\vspace{-3pt}
P(\Delta{\mathbf{p}}) = \sum_{k=1}^{K} \pi_k \mathcal{N}\left(\Delta{\mathbf{p}} \middle| \mu_{\mathbf{p},k}, \sigma_{\mathbf{p},k}^2 \right),
\end{equation}
where $P(\Delta{\mathbf{p}})$ represents the dispersion probability, $\mathbf{p}=[x;y]$ is the 2-D position, $\pi_k$ is the coefficient for the $k$-th Gaussian parameterized with mean $\mu_{\mathbf{p},k}$ and variance $\sigma_{\mathbf{p},k}^2$.

\begin{table*}[t]
\centering
\footnotesize
\begin{tabular}{l|p{0.815cm}<{\centering}p{0.815cm}<{\centering}|p{0.815cm}<{\centering}p{0.815cm}<{\centering}|p{0.815cm}<{\centering}p{0.815cm}<{\centering}|p{0.815cm}<{\centering}p{0.815cm}<{\centering}|p{0.815cm}<{\centering}p{0.815cm}<{\centering}|p{0.815cm}<{\centering}p{0.815cm}<{\centering}}
\toprule
\multicolumn{13}{c}{\textbf{Metalens Segmentation}} \\
\cmidrule(lr){1-13}
& 
\multicolumn{6}{c|}{\textbf{In-Vivo Diagnosis}} & 
\multicolumn{4}{c|}{\textbf{In-Vivo Surgery}} & 
\\ \cmidrule(lr){2-13}
& 
\multicolumn{2}{c|}{Meta-CVC-Clinic} & 
\multicolumn{2}{c|}{Meta-CVC-Colon} & 
\multicolumn{2}{c|}{Meta-Kvasir-Seg} & 
\multicolumn{2}{c|}{Meta-EndoVis-17} & 
\multicolumn{2}{c|}{Meta-EndoVis-18} &
\multicolumn{2}{c}{\textbf{Average}} \\ 
\cmidrule(lr){1-13}
Method & mIoU & mDICE & 
mIoU & mDICE & 
mIoU & mDICE & 
mIoU & mDICE & 
mIoU & mDICE &
mIoU & mDICE \\ 
\midrule
UNet~\cite{ronneberger2015u} & 0.6623 & 0.7710 & 0.6238 & 0.7306 & 0.4814 & 0.6171 & 0.7174 & 0.8274 & 0.5931 & 0.7181 & 0.6156 & 0.7328 \\
Atten-UNet \cite{oktay2018attention} & 0.6705 & 0.7752 & 0.6336 & 0.7306 & 0.4884 & 0.6255 & 0.7040 & 0.8163 & 0.5895 & 0.7200 & 0.6172 & 0.7335 \\
UNet++~\cite{zhou2018unet++} & 0.6765 & 0.7770 & 0.6417 & 0.7391 & 0.4746 & 0.6084 & 0.7195 & 0.8287 & 0.6003 & 0.7296 & 0.6225 & 0.7366 \\
UNext~\cite{valanarasu2022unext} & 0.7006 & 0.7948 & 0.6855 & 0.7707 & 0.5504 & 0.6651 & 0.7290 & 0.8377 & 0.5235 & 0.6496 & 0.6378 & 0.7436 \\
Rolling-UNet~\cite{liu2024rolling} & 0.7499 & 0.8259 & 0.6597 & 0.7460 & 0.5376 & 0.6482 & 0.7978 & 0.8847 & 0.6092 & 0.7359 & 0.6708 & 0.7681 \\
U-KAN~\cite{li2024u} & 0.7454 & 0.8260 & 0.7191 & 0.7924 & 0.5929 & 0.6937 & 0.7753 & 0.8692 & 0.6482 & 0.7766 & 0.6962 & 0.7916 \\
DeepLabV3+~\cite{chen2018encoder} & 0.6574 & 0.7506 & 0.6121 & 0.6943 & 0.4629 & 0.5882 & 0.7887 & 0.8784 & 0.6749 & 0.7956 & 0.6392 & 0.7414 \\
UperNet~\cite{xiao2018unified} & 0.6312 & 0.7201 & {0.7159} & 0.8041 & 0.5839 & 0.6910 & 0.8038 & 0.8810 & 0.6774 & 0.7999 & 0.6824 & 0.7792 \\
Segformer~\cite{xie2021segformer} & 0.7612 & 0.8378 & 0.6863 & 0.7729 & 0.4739 & 0.5970 & 0.8025 & 0.8878 & 0.6745 & 0.7965 & 0.6797 & 0.7784 \\
EDAformer~\cite{yu2025embedding} & \TopTwo{0.8169} & \TopTwo{0.8859} & \TopTwo{0.7507} & \TopTwo{0.8312} & 0.5705 & 0.6813 & 0.8161 & 0.8965 & \TopTwo{0.7310} & \TopTwo{0.8380}& 0.7370 & 0.8266 \\
SegNext~\cite{guo2022segnext} & 0.7639 & 0.8399 & 0.6528 & 0.7352 & 0.5790 & 0.6803 & 0.8207 & 0.8994 & 0.6266 & 0.7555 & 0.6886 & 0.7821 \\
Mask2former~\cite{cheng2021mask2former} & 0.7912 & 0.8612 & 0.7353 & 0.8157 & \TopTwo{0.6664} & \TopTwo{0.7558} & \TopTwo{0.8367} & \TopTwo{0.9010} & {0.7300} & {0.8371} & \TopTwo{0.7519} & \TopTwo{0.8342} \\
\midrule
MetaScope (Ours) & \best{0.8555} & \best{0.9137} & \best{0.7868} & \best{0.8604} & \best{0.6972} & \best{0.7845} &\best{0.8802}& \best{0.9355} & \best{0.8057} & \best{0.8896} & \best{0.8051} & \best{0.8767} \\ 
\midrule
\multicolumn{13}{c}{\textbf{Metalens Restoration}} \\
\cmidrule(lr){1-13}
Method & PSNR & SSIM & 
PSNR & SSIM & 
PSNR & SSIM & 
PSNR & SSIM & 
PSNR & SSIM &
PSNR & SSIM \\ 
\midrule
SwinIR \cite{liang2021swinir} & 25.3100 & 0.8442 & 24.1285 & 0.8338 & 23.7643 & 0.8328 & 26.1958 & 0.9143 & 25.2748 & 0.8786 & 24.9347 & 0.8607 \\
Restormer \cite{zamir2022restormer} & 33.5640 & 0.9581 & 33.9541 & \TopTwo{0.9719} & 31.3738 & 0.9778 & 30.9982 & 0.9561 & 30.6164 & 0.9356 & 32.1013 & 0.9599 \\
NAFNet \cite{chen2022simple} & \TopTwo{33.8970} & 0.9668 & 34.6834 & \best{0.9856} & 31.5939 & \TopTwo{0.9793} & 30.9660 & 0.9553 & 30.9087 & 0.9397 & 32.4098 & \TopTwo{0.9653} \\
DRMI \cite{seo2023deep} & 33.5517 & \TopTwo{0.9784} & \TopTwo{34.7755} & 0.9672 & \TopTwo{31.6620} & 0.9705 & \TopTwo{31.6709} & \TopTwo{0.9601} & \TopTwo{31.4767} & \TopTwo{0.9451} & \TopTwo{32.6274} & 0.9643 \\
MambaIR \cite{guo2025mambair} & 25.5517 & 0.8689 & 26.9477 & 0.9128 & 24.7844 & 0.9165 & 24.1641 & 0.8852 & 24.3246 & 0.8572 & 25.1545 & 0.8881 \\
NeRD-Rain~\cite{chen2024bidirectional} & 30.3351 & 0.9370 & 31.1423 & 0.9558 & 28.3791 & 0.9564 & 25.7874 & 0.9072 & 25.4942 & 0.8712 & 28.2276 & 0.9255 \\
\midrule
MetaScope (Ours) & \best{34.8483} & \best{0.9854} & \best{35.6603} & \best{0.9856} & \best{32.4454} & \best{0.9818} & \best{32.0365} & \best{0.9619} & \best{31.8664} & \best{0.9474}& \best{33.3714} & \best{0.9724} \\
\bottomrule
\end{tabular}\vspace{-2pt}
\caption{Comparison with state-of-the-art methods on five benchmarks in terms of metalens segmentation (top) and restoration (bottom). The best and second-best results are highlighted in red and blue, respectively.}
\label{tab:sota_reorganized_en}\vspace{-8pt}
\end{table*}

\noindent\textbf{Learning PSF Distribution via Multi-Expert Gaussian.} Based on the GMM formulation, we aim to learn a spatial deformation $\Delta{\mathbf{p}}$ for each position $\mathbf{p}$ to model PSFs, adhering $P(\Delta{\mathbf{p}})$. This can thereby facilitate the adaptive aggregation of dispersed information. However, directly estimating the GMM in Eq.~\ref{eqn_gmm} with conventional methods like the EM algorithm is time-consuming and challenging since each wavelength $\lambda$ is associated with a specific GMM with diverse patterns. To break through this barrier, inspired by the Mixture of Experts (MoE)~\cite{masoudnia2014mixture}, we propose a Multi-expert Gaussian (MeG) paradigm with $K$ Gaussian experts, where each expert $\text{MeG}_k$ serves for a specific Gaussian pattern. Notably, instead of directly estimating the Gaussian parameters, the expert is designed to estimate the linear transformation from a normal Gaussian, i.e., \emph{residual} part, based on the scaling invariance of the Gaussian.

Specifically, given the input feature $X$, we first assume a normally distributed latent $P(z|X)$ with variational autoencoding for the PSF distributed offset space. To this end, we deploy two linear layers for the mean $\boldsymbol{\mu}({X})$ and log-variance $\log \boldsymbol{\sigma}^2({X})$ as Gaussian parameters and implement Kullback-Leibler divergence to ensure that offset space to be Gaussian: $\text{KL}\left(q_\phi(\boldsymbol{z} \mid X) \| \mathcal{N}(0, \mathbf{I})\right)$ as~\cite{kingma2013auto}, where $q_\phi$ is simply formulated with the former $\textit{Conv}$ block~\cite{li2024cliff}. Then, we establish $K$ experts by leveraging the linear invariance property of Gaussian distributions, where each of the experts parameterized with $\{\mathbf{a}_{k}, \mathbf{b}_{k} \}$ aims to learn a linear projection based on a sampled $z \sim P(z|X)$ using the reparameterization trick~\cite{kingma2013auto}. Thus, this can generate a transformed Gaussian for each expert: $\text{MoG}_{k}(X) \sim \mathcal{N} (\mathbf{a}_{k}\boldsymbol{\mu}({X}) + \mathbf{b}_{k}, \mathbf{a}_{k}^2\boldsymbol{\sigma}^2({X}))$. Concretely, each $\text{MeG}$ is deployed with a convolution layer outputting $2\times M\times M$ channels where $M=3$, which generates $M^2$ offset candidates $[\Delta x; \Delta y]$ modeling the implicit PSFs\footnote{To avoid the collapse to the same Gaussian and facilitate learning orthogonally, we deploy the MeG layers with different dilation rates.}. Finally, a small router layer $\text{Router}(X)$ is deployed with $\texttt{Conv}- \texttt{Pooling}$ to learn the aggregation weight with $K$-channel output. Thus, the whole process can be formally represented as, 
\begin{equation}\label{eq:offset}
\Delta{\mathbf{p}} = \sum_{k=1}^{K} \pi_k \cdot \text{MeG}_k(X), \text{where } \pi = \text{Router}(X),
\end{equation}
and $ \Delta{\mathbf{p}}$ is the estimated spatial offsets in modeling dispersion patterns. These offsets are utilized to aggregate dispersed points with the well-estimated guidance of PSFs.

\noindent\textbf{Chromatic Correction.} With the learned offsets, we further aggregate the spatially dispersed points with a simple deformable convolution layer, facilitating local aggregation of dispersed information. This can ensure that the pixel colors are accurately reconstructed by aligning with the underlying physical distribution modeled by the GMM, denoted as ${X}(\textbf{p})_{\text{OCC}} = \sum_{i=1}^{M+N} w_i \cdot X(\mathbf{p} + \Delta \mathbf{p}_{i})$, where $\Delta \mathbf{p}_i$ represents the learned offset for the $i$-th element ($i \in [1, M\times M]$) at location $\mathbf{p}$, and $w_i$ is the learnable weight. By learning the positions of the sampled points, we can effectively mitigate color dispersion caused by PSFs, leading to more accurate and visually consistent color correction.

\subsection{Optimization}
\noindent\textbf{Gradient-guided Distillation.} To better optimize the proposed optics-informed modules in chromatic aberration learning, we propose a distillation method that leverages the pre-trained DINO v2~\cite{oquab2023dinov2} to refine under-optimized regions of $\Tilde{X}$, as illustrated in Fig.~\ref{fig:overall}c. Given the feature $\Tilde{X}$, we back-propagate the restoration Loss $L_{\text{Rest}}$ and collect the gradient amplitude map $\mathbf{M}=|\frac{\partial L_{\text{Rest}}}{\partial \tilde{X}}|$ to guide the feature distillation, which can be denoted as follows,
\begin{equation}
L_{\text{Distill}} = \frac{1}{\text{sum}(\Tilde{\mathbf{M}})} \sum \Tilde{\mathbf{M}} \odot \left| \mathcal{F}_{\text{DINO}}({I}_c) - \mathcal{F}^{\text{Proj}}_{\text{Conv}}(\tilde{X}) \right|_1,
\end{equation}
where $\Tilde{\mathbf{M}}=\frac{\mathbf{M} - \min(\mathbf{M})}{\max(\mathbf{M}) - \min(\mathbf{M})}$ is the normalized gradient map, $\mathcal{F}^{\text{Proj}}_{\text{Conv}}$ is a projection layer structured as $\texttt{Conv}- \texttt{ReLU}- \texttt{LayerNorm}-\texttt{Conv}$ to match the dimension. Consequently, the mask with higher values corresponds to areas with greater distortion and under-optimization, enabling the distillation to focus preferentially on critical regions.

\noindent\textbf{Overall Objective.} During the training stage of MetaScope, the overall optimization objective is denoted as:
\begin{equation}\label{eq:all_loss}
\mathcal{L}_{\text{MetaScope}} = L_{\text{Seg}} +\lambda L_{\text{Rest}} + \omega_d L_{\text{Distill}} + \omega_k L_{\text{KL}},
\end{equation}
where $L_{\text{Rest}}$ is for the auxiliary metalens restoration task, deployed with PSNR loss, $ L_{\text{Seg}}$ is the segmentation loss, implemented with CE loss, $L_{\text{KL}}$ is the variational loss to ensure the Gaussian~\cite{kingma2013auto} in OCC estimating PSF distributions. $\omega_d$ and $\omega_k$ balance the two representation learning modules. $\lambda$ is used to balance the segmentation and restoration.

\section{Experiments}

\subsection{Benchmark Setup} \label{sec:benchmark}

\noindent{\textbf{Dataset Setup.}} We build up five benchmarks focusing on two critical clinical procedures: diagnosis and surgery. After establishing the metalens optical system, as described in Sec.~\ref{sec:dataset}, we are able to capture the metalens image $I_{m}$ by photographing the publicly available database $\{I_{c}, Y\}$ projected on the screen, thereby generating the triplet data: $\{I_{m}, I_{c}, Y\}$ in $1024 \times 1024$ resolution. Based on our dataset, we carry out two streams of research to advance the metalens vision, including metalens segmentation with $\{I_{m}, Y\}$ (perception), metalens restoration $\{I_{m}, I_{c}\}$ (low-level vision). Hence, we name our datasets by adding the \textbf{Meta} prefix on the following five database: {(1) CVC-Clinic}~\cite{bernal2015wm}, {(2) CVC-Colon}~\cite{bernal2012towards} {(3) Kvasir-Seg}~\cite{jha2020kvasir}, {(4) EndoVis-17}~\cite{allan20192017}, and {(5) EndoVis-18}~\cite{allan20202018}, such as Meta-EndoVis-18. More details can be found in Supp. D.

\noindent{\textbf{Evaluation Metrics.}} Evaluation is conducted in alignment with two clinical needs. (1) Clinical assistance: Metalens segmentation adopts mIoU and mDICE~\cite{li2024u} (IoU/DICE for single-class datasets). (2) Clinical visualization: Metalens restoration uses PSNR and SSIM~\cite{zamir2022restormer}.

\noindent{\textbf{Implementation Details.}} The proposed model is based on the NAFNet~\cite{chen2022simple} and is trained using the AdamW optimizer with a learning rate of $1 \times 10^{-3}$, a batch size of 2, and a weight decay of $1 \times 10^{-4}$. The training schedules of all benchmarked methods are consistent. Training was conducted on two 24G GPUs for 50,000 iterations. The number of MeG is set to $K = 3$. The $\omega_d$ and the $\omega_k$ are both set to 1, and the $\lambda$ is set to 0.01/0.1 for diagnostics/surgery datasets. The simulation details and results of the optical priors in Eq.~\ref{eq:optical embedding} are in Supp. B. 

\subsection{Comparison with State-of-the-art Methods}

\begin{figure}[t]
\centering
\includegraphics[width=1.0\linewidth]{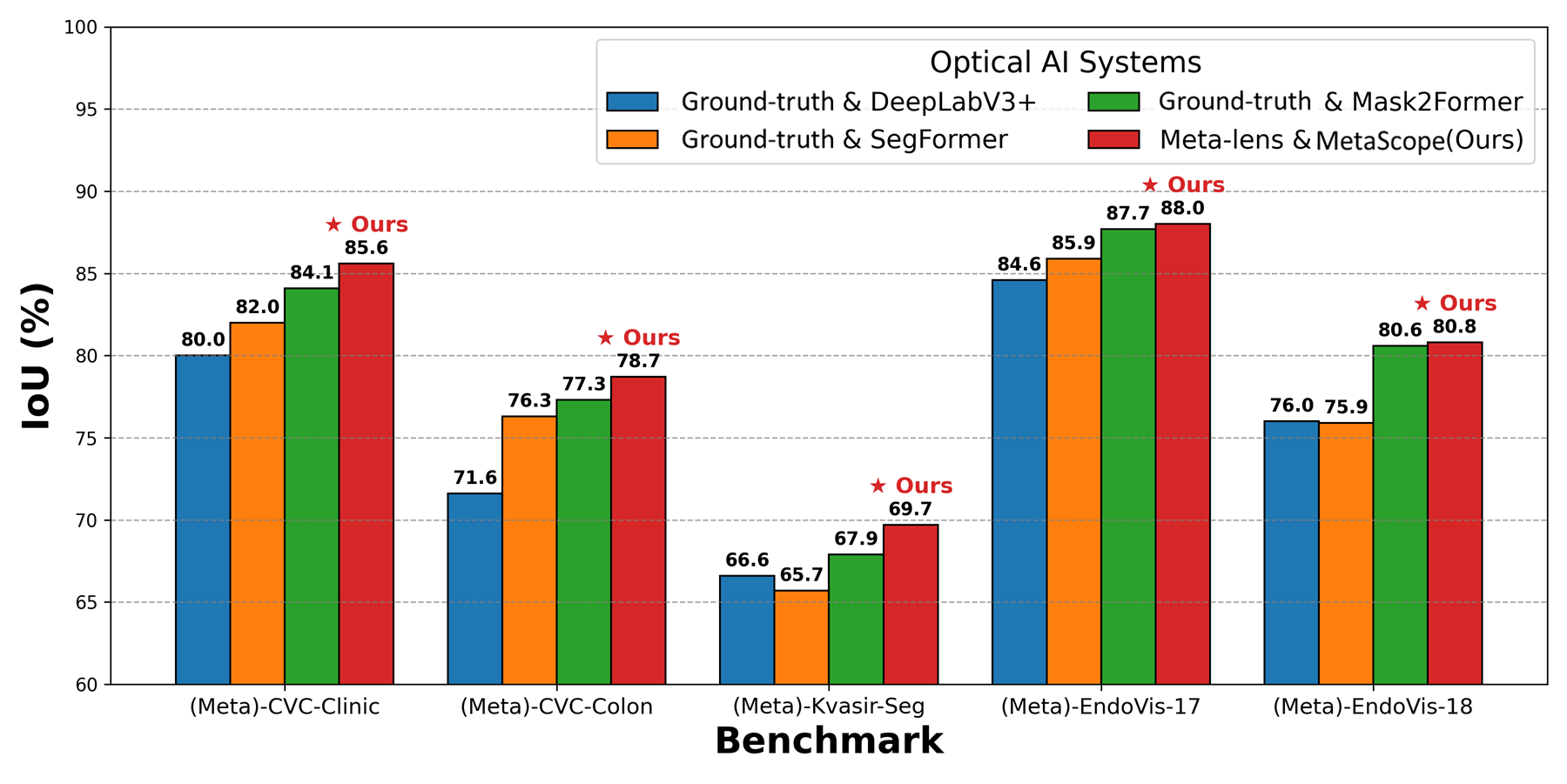}\vspace{-10pt}
\caption{Comparison between conventional systems using high-quality ground-truth images without \textbf{Meta}, and our metalens system using the degraded images in \textbf{Meta} version dataset.}\label{fig:MetaVSConvex}
\vspace{-10pt}
\end{figure}

\begin{table}[t]
\footnotesize
\centering
\begin{tabular}{l|c|cc|cc}
\toprule
\multicolumn{2}{c|}{Setting} & \multicolumn{2}{c|}{Segmentation} & \multicolumn{2}{c}{Restoration} \\
\midrule
- & w/o & mIoU & mDICE & PSNR & SSIM \\
\midrule
Seg-Base& -& 0.7376 & 0.8272 & - & - \\
Meta-Base & -& 0.7749 & 0.8573 & 31.1748 & 0.9538 \\
\midrule
\multirow{2}{*}{+ Grad-Distill} & Grad & 0.7975&0.8750 &32.4870 & 0.9632\\
& - &0.8312 & 0.8927	 &32.8153 & 0.9610\\\midrule
{+ Grad-Distill;} & MeG & 0.8382	& 0.9012 & 32.8078& 0.9482 \\
{+ OCC} & - & 0.8512& 0.9088	&34.0011 & 0.9603 \\\midrule
{+ Grad-Distill;} & Prior &0.8523 & 0.9109 &34.1341 &0.9806 \\
{+ OCC; + OIA} & - &\bestSmall{0.8555} &\bestSmall{0.9137} & \bestSmall{34.8483} & \bestSmall{0.9854} \\ \bottomrule
\end{tabular}
\caption{Detailed ablation study of the proposed modules on the Meta-CVC-Clinic benchmark, where `w/o' indicates removing the sub-components, and `Grad' indicates the gradient guidance.}
\label{tab:ablation}
\end{table}

\begin{table}[t]
\footnotesize
\centering
\begin{tabular}{c|c|cc|cc}
\toprule
Module & Setting & mIoU & mDICE & PSNR & SSIM \\ \midrule
& Learned & {0.8700}	 & {0.9293} & 31.6896 & 0.9600	\\
\multirow{1}{*}{Optical} & Only ch. &{0.8715} &{0.9303} & {31.8751} & 0.9615 \\
\multirow{1}{*}{Prior}& Only sp. &{0.8746} &{0.9320} & {32.0141} & {0.9612} \\
& Full &\bestSmall{0.8802} &\bestSmall{0.9355} & \bestSmall{32.0365} & \bestSmall{0.9619} \\
\midrule
\multirow{2}{*}{Num. MeG } &1 & 0.8682	 & 0.9279 & 31.8780 & 0.9616 \\
\multirow{2}{*}{$K$ in Eq.~\ref{eq:offset} }& 3 &\bestSmall{0.8802} &\bestSmall{0.9355} & {32.0365} & {0.9619} \\
&5 &{0.8793} &{0.9349} & \bestSmall{32.0637} & \bestSmall{0.9625} \\ 
\midrule
\multirow{2}{*}{Loss Weight} &0.01 &0.8676 &{0.9280} & {29.8646} & {0.9452} \\
\multirow{2}{*}{ $\lambda$ in Eq.~\ref{eq:all_loss}} &0.1&\bestSmall{0.8802} &\bestSmall{0.9355} & {32.0365} & {0.9619} \\
&1.0 &{0.8497} &{0.9168} & \bestSmall{32.0491} & \bestSmall{0.9622} \\
\bottomrule
\end{tabular}
\caption{Sensitivity analysis of the simulated optical prior and hyper-parameters on the Meta-EndoVis-17 benchmark, where `Learned' indicates directly learning optical embedding.}
\label{tab:further_analysis}
\end{table}

\begin{figure*}[t]
\centering
\includegraphics[width=0.99\linewidth]{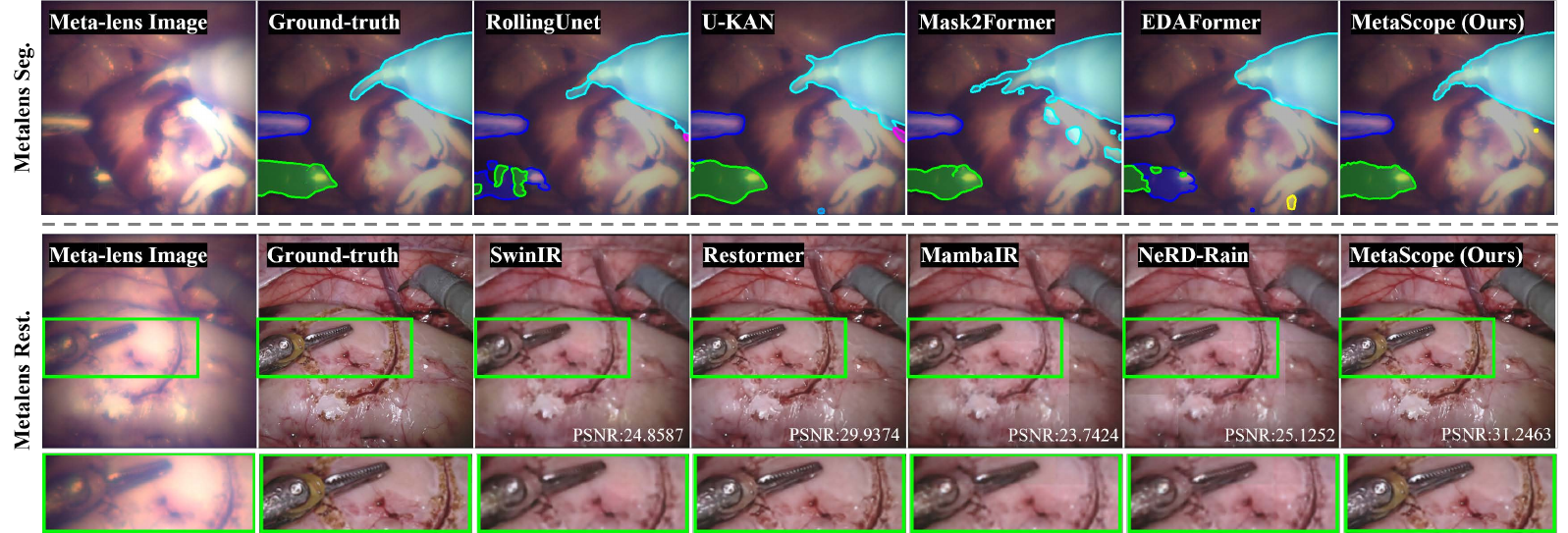} 
\vspace{-5pt}
\caption{Qualitative comparison in terms of state-of-the-art segmentation methods and image restoration methods.}\label{fig:show}
\vspace{-15pt}
\end{figure*}

\noindent{\textbf{Metalens Segmentation.}} In Tab.~\ref{tab:sota_reorganized_en} (top), we compare with state-of-the-art methods in both the medical field like U-KAN~\cite{li2024u} and Rolling-UNet~\cite{liu2024rolling} and the natural imaging, including Mask2former~\cite{cheng2021mask2former}. Our MetaScope achieves superior performance, attaining an 80.51\% mIoU and 87.67\% mDICE across all five benchmarks, significantly outperforming the latest counterpart, EDAformer~\cite{yu2025embedding}, by 6.81\% in mIoU and 5.01\% in mDICE. Most importantly, MetaScope exhibits the best performance across all five settings, encompassing varied clinical devices, in-vivo scenes, and diverse classes, demonstrating substantial robustness.

\noindent{\textbf{Metalens Restoration.}} Given promising achromatic metalenses~\cite{lin2019achromatic}, we benchmark metalens restoration and compare our method with state-of-the-arts, as shown in Tab.~\ref{tab:sota_reorganized_en} (bottom). MetaScope achieves the best performance with a PSNR of 33.3714 dB and an SSIM of 0.9724, surpassing the latest metalens restoration method DRMI~\cite{seo2023deep} and the natural imaging method MambaIR~\cite{guo2025mambair}, revealing that our method can thoroughly correct distorted patterns. We also observe that methods tailored for specific restoration, such as NeRD-Rain for de-raining, perform unsatisfactorily, highlighting the unique characteristics of metalens (Sec.~\ref{sec:preliminaries}).

\noindent{\textbf{Comparing Metalens and Conventional Systems.}} Fig.~\ref{fig:MetaVSConvex} presents a more challenging comparison between metalens and conventional systems, where the latter utilizes high-quality ground-truth images and state-of-the-art segmentation algorithms. Surprisingly, despite metalens introducing chromatic aberration, our system ({red bar}) achieves the best performance across all benchmarks. This demonstrates that our method is robust to chromatic aberration and capable of unleashing the potential of metalens. \emph{Notably, our system transcends the physical weight limitations and algorithmic precision constraints of conventional lens systems.}

\subsection{Ablation Study}

\noindent{\textbf{Effect of Each Module.}} We conduct a comprehensive ablation study presented in Tab.~\ref{tab:ablation} and summarize several valuable observations. (1) Compared with the pure baseline (Seg-Base), introducing the restoration branch (Meta-Base) to guide segmentation yields a noticeable 3.01\% mDICE gain. (2) The gradient-guided distillation contributes an additional 3.54\% mDICE enhancement, highlighting the critical role of the restoration gradient (Grad). (3) Introducing OCC enhances the mDICE by 1.61\% and achieves a 1.18 dB gain in PSNR, with the restoration gains primarily arising from the MeG design, emphasizing the importance of modeling dispersion patterns. (4) Incorporating OIA further refines the restoration representation, achieving the best performance in both segmentation (91.37\% mDICE) and restoration (34.8483 PSNR), demonstrating the effectiveness of the optical prior in meta-restoration.

\begin{figure}[t]
\centering
\includegraphics[width=1.0\linewidth]{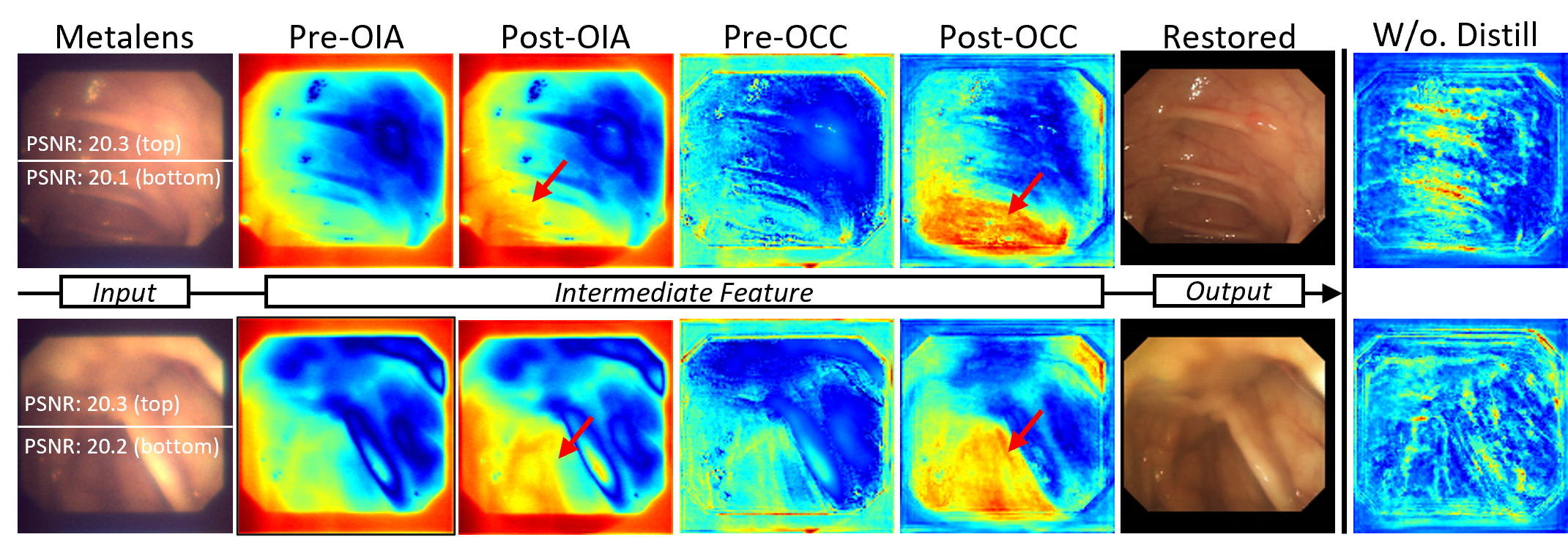}
\vspace{-15pt}
\caption{Heatmaps of the feature before and after each module.}\label{fig:feat}
\vspace{-20pt}
\end{figure}

\noindent{\textbf{Sensitivity Analysis.}} We further conduct in-depth exploration in Tab.~\ref{tab:further_analysis}. (1) Removing the spatial and channel priors results in a noticeable decrease in PSNR, highlighting the essential role of correcting intensity attenuation. (2) Regarding the number of MeG, the performance remains robust when $K \geq 3$ but experiences a noticeable decrease when $K=1$, revealing the effectiveness of modeling PSFs by the Gaussian mixture. (3) There is a trade-off in joint learning: Increasing the restoration loss weight yields further gains in PSNR while sacrificing some mDICE.

\subsection{Qualitative Analysis} \label{sec:qualitative}

\textbf{Qualitative Comparison.} Fig.~\ref{fig:show} shows a comparison between segmentation (top) and restoration methods (bottom). Compared to the latest method, EDAFormer~\cite{yu2025embedding}, MetaScope shows visually significant improvements in the spatial completeness of challenging local regions. This is particularly evident in the prograsp forceps (green) and monopolar curved scissors (cyan) in row 1. Regarding the metalens restoration, Metascope achieves significant gains in both color accuracy and structural fidelity of local regions.

\noindent\textbf{Analyze each Module with Heatmaps.} In Fig.~\ref{fig:feat}, we progressively visualize features before and after each module, highlighting \emph{\textbf{insightful interoperability and well-aligned optical motivation}}. \textbf{(1)} By modulating features with spatial priors, OIA enhances the near-border regions and relieves the position-dependent intensity loss, showing a similar pattern to Fig.~\ref{fig:two issues}c. \textbf{(2)} OCC encourages each pixel to cluster with its surrounding pixels, addressing PSF-caused dispersion with a spatial grouping (consistent with Fig.~\ref{fig:two issues}d). It also focuses on more distorted regions (bottom part) instead of segmenting targets, perfectly aligning with chromatic correction designs. \textbf{(3)} Without distillation, only low-level details are focused, yet lacking semantic guidance.

\section{Conclusion}

For the first attempt, we study endoscopic intelligence on the microscale with advanced metalenses. We establish benchmarks with five datasets across diagnostic and surgical applications. To address the chromatic aberration, we propose a novel MetaScope with two key components: \textit{Optical-informed Intensity Adjustment} corrects intensity attenuation by learning optical embeddings; \textit{Optical-informed Chromatic Correction} models and eliminates PSF distortion with multi-expert Gaussian. A gradient-guided distillation is integrated to enhance the meta-representation. Extensive experiments demonstrate that MetaScope significantly surpasses state-of-the-art methods.

\section*{Acknowledgement}

This research was partly supported by Hong Kong PhD Fellowship from Research Grants Council of Hong Kong. This work was supported by Innovation and Technology Fund ITS/229/22, PRP/082/24FX and Hong Kong Research Grants Council (RGC) General Research Fund 14204321. We acknowledge the support from the University Grants Committee / Research Grants Council of the Hong Kong Special Administrative Region, China [Project No. AoE/P-502/20, CRF Project C5031-22G, GRF Project: CityU11305223; CityU11300224, CityU11310522, CityU11300123], National Natural Science Foundation of China [Grant No. 62375232], Guangdong Basic and Applied Basic Research Foundation [2025A1515011846], City University of Hong Kong [Project No. 9380131, 9610628] and the Shenzhen Research Institute, City University of Hong Kong [R-IND27901].

{
\small
\bibliographystyle{ieeenat_fullname}
\bibliography{main}
}

\appendix
\clearpage
\setcounter{page}{1}
\maketitlesupplementary
The supplementary is organized as follows.
\begin{enumerate}[label={}]
\item Supp.~\ref{background}: Details of metalens:
\begin{itemize}
\item metalens imaging,
\item metalens fabrication,
\item chromatic aberration,
\item existing physical and algorithmic solutions.
\end{itemize}
\item Supp.~\ref{optical}: Optical simulation details of:
\begin{itemize}
\item spatial optical prior,
\item channel optical prior.
\end{itemize}
\item Supp.~\ref{cv}: Experiments including:
\begin{itemize}
\item {real scene generalization}
\item data scaling-up,
\item cross-metalens generalization,
\item model efficiency,
\item dataset analysis,
\item extensive qualitative analysis.
\end{itemize}
\item Supp.~\ref{discussion}: Discussions and clarifications about:
\begin{itemize}
\item dataset setup,
\item physically suitable for endoscopy,
\item related works,
\item mathematical proofs,
\item ethical clarification.
\end{itemize}
\end{enumerate}

\section{Delving into Metalens}\label{background}

Traditional convex lenses focus light by varying the thickness from the center to the edge, altering the optical path length of incoming light. However, this design inherently causes spherical and chromatic aberrations. Chromatic aberration is further divided into axial chromatic aberration (ACA) and lateral chromatic aberration (LCA). ACA involves variations in focal length along the optical ($z$) axis for different wavelengths. LCA results in the positional displacement of focused colors in the transverse ($x$-$y$) plane on the focal plane. Conventional convex lenses employ multiple lens elements arranged in groups to mitigate these aberrations, often leading to bulky optical systems.

Differently, metalenses~\cite{khorasaninejad2016metalenses} \emph{(published in Science, 2016)} represent a novel class of lenses that are ultra-lightweight and free from bulky designs. These lenses comprise numerous sub-wavelength structures (scale elements within the visible spectrum) meticulously arranged on a planar surface. Each nanostructure can independently manipulate the phase of the transmitted light wave, allowing for precise deformation of the wavefront and achieving accurate focusing. By programming a target hyperbolic phase profile that inherently satisfies the Abbe sine condition, metalenses can theoretically eliminate spherical aberration at the design stage. Combined with their sub-micron thickness and planar form factor, this paradigm shift allows metalenses to achieve high-performance focusing without the bulk of traditional optics, offering transformative potential for miniaturized imaging systems and photonic integration.

\subsection{Metalens Design and Fabrication} 

However, optimizing the achromatic aberration of metalenses remains challenging in meta-optics. The focusing phase of a metalens for a specific working wavelength $\lambda$ is determined by the following equation:
\begin{equation}\label{eq:focusing}
\phi(r, \lambda) = -\frac{2\pi}{\lambda} \left( \sqrt{r^2 + f^2} - f \right), \quad \lambda \in {[\lambda_{\text{min}}, \lambda_{\text{max}}]}
\end{equation}
where $\phi(r, \lambda)$ represents the required focusing phase at position $r$ for the wavelength $\lambda$, and $f$ denotes the desired focal length. In this study, the metalens, with a diameter of 2.6 mm, is designed to achieve a focal length of 10 mm at a wavelength of 532 nm.

\begin{figure}[t]
\begin{center}
\includegraphics[width=1.0\linewidth]{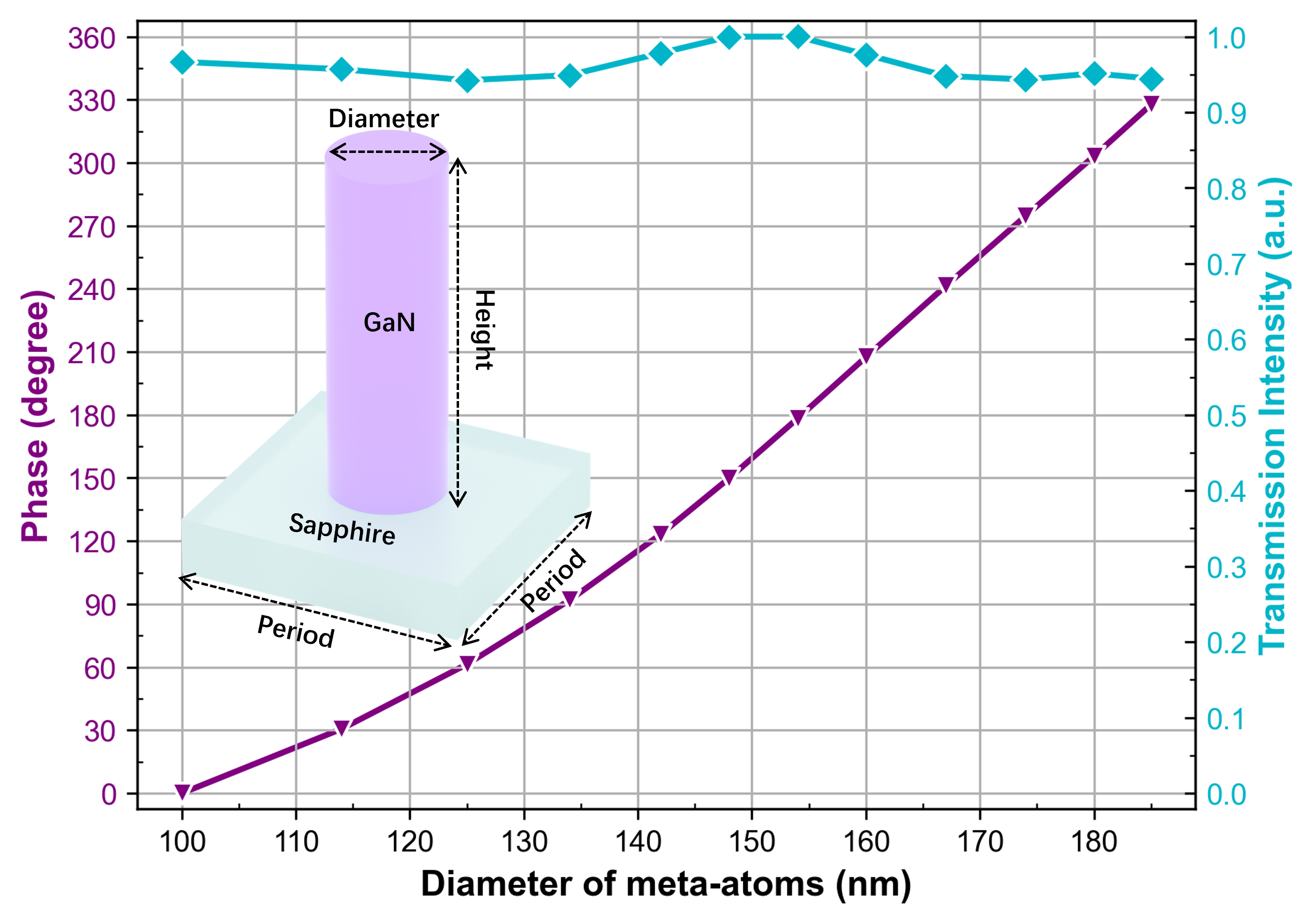} 
\caption{Design of meta-atoms. The phase and transmission intensity of the GaN nanopillars. }\label{fig:mod} \vspace{-20pt}
\end{center}
\end{figure}

Fig.~\ref{fig:mod} shows the employed polarization-independent meta-atoms, comprising cylindrical gallium nitride (GaN) nanopillars on a sapphire substrate. Each meta-atom features a fixed height of 850 nm and a unit-cell periodicity of 280 nm. The parametric variation of the nanopillar diameters (100–185 nm) enables full $2\pi$-phase coverage for wavefront focusing. The data of phase shift and transmission intensity are derived from numerical simulation with the commercial software COMSOL Multiphysics\copyright. As shown in Eq.~\ref{eq:focusing}, different meta-atoms are arranged at corresponding locations according to the focusing phase profile.

\begin{figure*}[t]
\begin{center}
\includegraphics[width=0.8\linewidth]{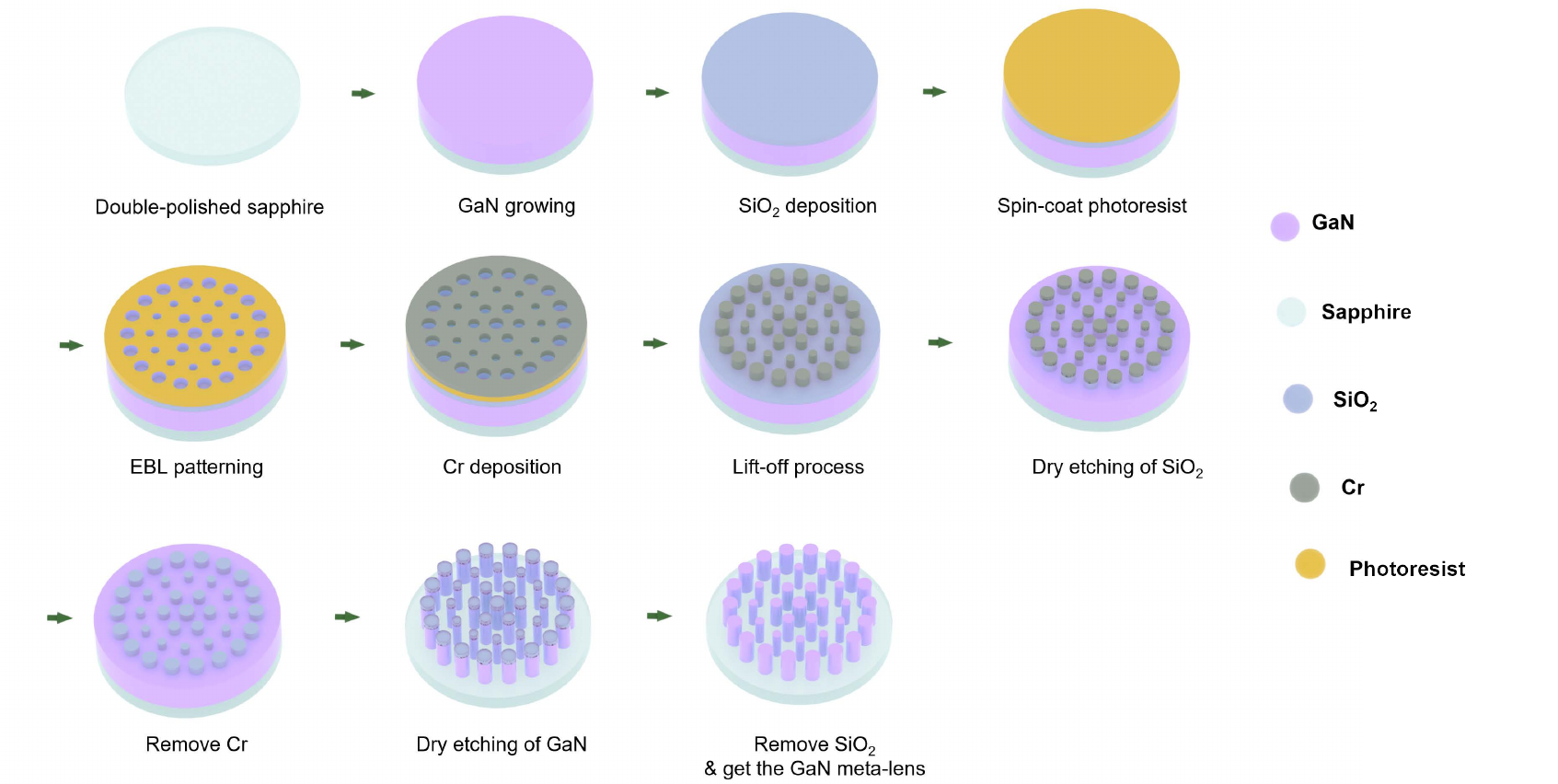} 
\caption{Fabrication flow chart of the metalens. }\label{fig:fab}
\end{center}\vspace{-15pt}
\end{figure*}
In Fig.~\ref{fig:fab}, the metalens is manufactured through a multi-step lithographic patterning and dry etching workflow. The process begins with an ultrasmooth c-plane sapphire substrate coated with an 850 nm GaN epitaxial layer grown via metalorganic chemical vapor deposition (MOCVD). A 200 nm silicon dioxide (SiO$_2$) film, serving as an etch-resistant hard mask, is thermally evaporated onto the GaN surface. To pattern the metasurface, a polymethyl methacrylate (PMMA) photoresist layer (200 nm thick) is deposited via spin coating, followed by a 180°C soft bake for 3 minutes. High-resolution electron-beam lithography (ELS-HS50, ELIONIX INC.) directly writes the metalens design into the PMMA layer. Post-exposure development involves immersing the substrate in a methyl isobutyl ketone/isopropyl alcohol (MIBK: IPA = 1:3) solution for 75 seconds, followed by an IPA rinse (20 seconds) to terminate the reaction. A 40 nm chromium (Cr) film is subsequently evaporated onto the patterned resist, and lift-off in acetone selectively removes excess Cr to define the mask geometry. This Cr stencil guides the first inductively coupled plasma reactive ion etching (ICP-RIE) step using CF$_4$ plasma (Samco RIE-200iPT) to transfer the pattern into the underlying SiO$_2$ layer. Residual Cr is stripped via wet etching, exposing the SiO$_2$ hard mask. The final nanostructuring involves a second ICP-RIE cycle with a Cl$_2$/Ar plasma to anisotropically etch the GaN layer, resulting in high-aspect-ratio nanopillars. Buffered oxide etch (BOE) solution removes the remaining SiO$_2$ mask, leaving an array of precisely defined GaN nanostructures anchored to the sapphire substrate.

\subsection{Chromatic Aberrations} 

\begin{figure}[h]
\begin{center}
\includegraphics[width=0.8\linewidth]{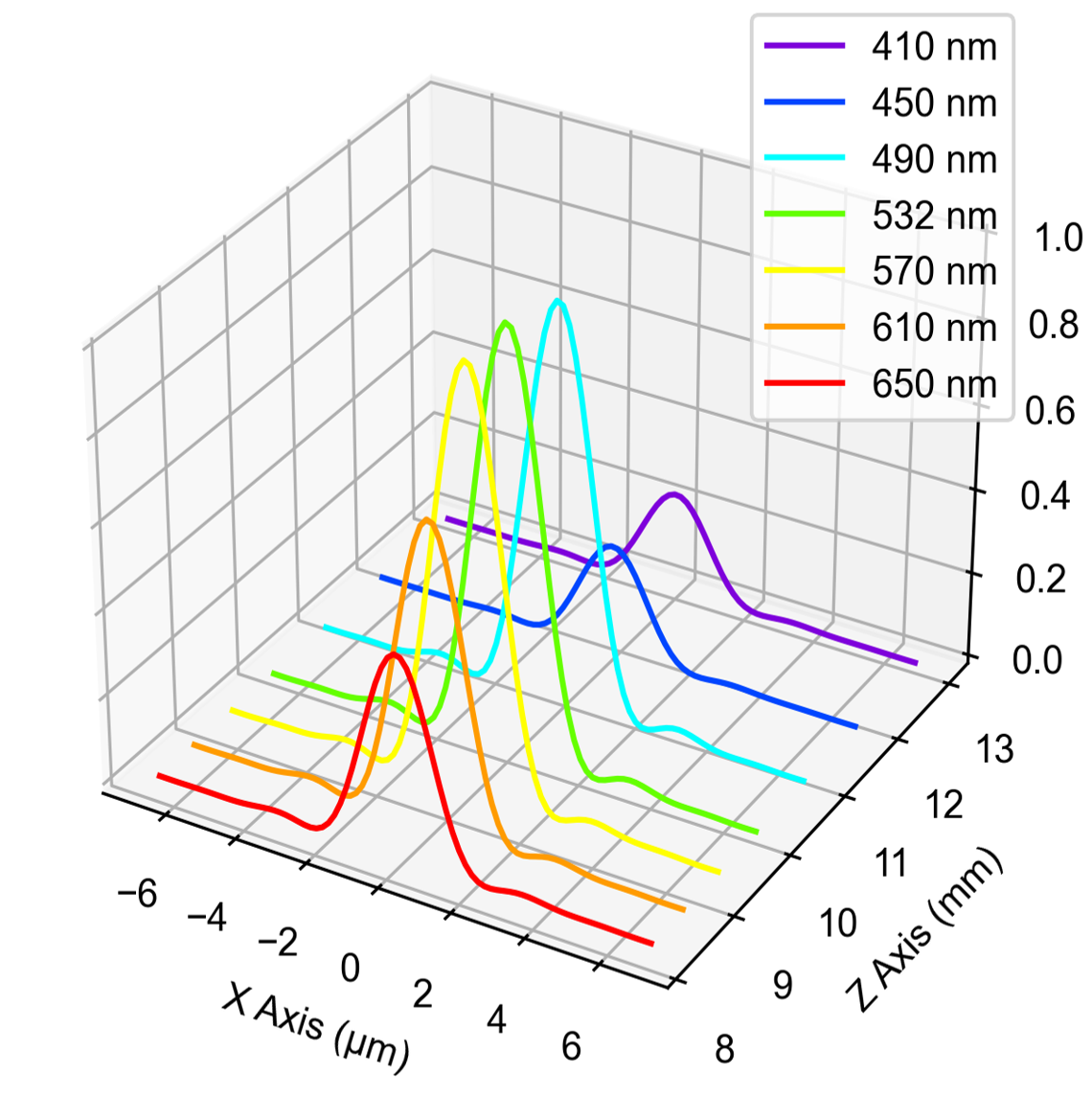} 
\caption{Focal lengths of the metalens for different wavelengths. }\label{fig:defocus}\vspace{-20pt}
\end{center}
\end{figure}

The focal length at a different wavelength can be estimated by scaling the designed focal length of 532 nm proportionally to the wavelength ratio, assuming that the lens material and structure introduce purely dispersive effects without significant aberrations~\cite{tseng2018metalenses,khorasaninejad2016metalenses}: $f(\lambda) = f_0 \times \frac{\lambda}{\lambda_0}$, where $f_0$ and $\lambda_0$ are the constants of the designed metalens parameters, facilitating the prediction of focal lengths across different wavelengths. As illustrated in Fig.~\ref{fig:defocus}, the light of different wavelengths (i.e., different colors) will have different focal lengths (along the $z$-axis). Red light has a shorter focal length, while blue light has a longer one. Due to these varying focal lengths, specific light colors will appear out of focus when capturing color images, resulting in blurring differently in each color channel. Mismatched color offsets can also create color fringing around objects, especially at high-contrast edges.

\label{tab:ch_prior}
\begin{table*}[h!]
\centering
\begin{tabular}{|>{\centering\arraybackslash}m{2cm}|*{7}{>{\centering\arraybackslash}m{1.5cm}|}}
\hline
Color & \cellcolor{red!50}Red & \cellcolor{orange!50}Orange & \cellcolor{yellow!50}Yellow & \cellcolor{green!50}Green & \cellcolor{cyan!50}Cyan & \cellcolor{blue!50}Blue & \cellcolor{violet!50}Violet \\
\hline
 Wavelength & 650 nm & 610 nm & 570 nm & 532 nm & 490 nm & 450 nm & 410 nm \\
\hline
Efficiency & 0.3524 & 0.5281 & 0.7371 & 0.9920 & 0.7032 & 0.1885 & 0.1738 \\
\hline
\end{tabular}
\caption{Comparison of the intensity at the focusing points for each color, serving as the transformation efficiency $\mathbf{T}$ in the proposed Optics-informed Intensity Adjustment (OIA) module. }
\label{tab:color_wavelength_efficiency}
\end{table*}
\subsection{How to Solve Metalens Chromatic Aberrations}

\noindent\textbf{Physical Solution.} To physically correct ACA by ensuring identical focal lengths across different wavelengths, the design of an achromatic metalens must incorporate an additional wavelength-dependent phase delay $\Delta\phi(r, \lambda)$ to achieve the optical rectification~\cite{dong2024achromatic,lin2019achromatic,wang2018broadband}:
\begin{equation}
\phi_{Achromatic}(r, \lambda) = \phi(r, \lambda_{max})+\Delta\phi(r, \lambda),
\end{equation}
\begin{align}
\Delta\phi(r, \lambda) &= -\left[2\pi\left(\sqrt{r^2 + f^2} - f\right)\right]\left(\frac{1}{\lambda} - \frac{1}{\lambda_{\text{max}}}\right) \notag \\
&\quad + \frac{\delta}{\lambda} \cdot \frac{\lambda_{\text{min}} \lambda_{\text{max}}}{\lambda_{\text{max}} - \lambda_{\text{min}}} - \frac{\delta \lambda_{\text{min}}}{\lambda_{\text{max}} - \lambda_{\text{min}}}.
\end{align}
Here, $\phi_{\text{Achromatic}}(r, \lambda)$ is the achromatic focusing phase at position $r$ for wavelength $\lambda$, and $\delta$ represents the maximum additional phase shift required. However, as the size of the metalens increases, the necessary additional phase delay also grows, posing significant and open challenges. \emph{Due to the limitations of current micro-nano processing technology, finding a solution solely through the geometric design of the nano-antennas proves challenging.}

\noindent\textbf{Computer Vision Solution.} To overcome inherent physical constraints, alternative approaches that leverage computer vision and computational optics have garnered significant attention for addressing ACA in metalenses~\cite{tseng2021neural,ueno2024ai,seo2023deep,zhelyeznyakov2021deep,chen2022artificial}. Recent advancements and the democratization of computing platforms have positioned these methods as promising solutions, effectively bridging both hardware and software limitations. By accurately modeling the degradation patterns caused by ACA, machine learning algorithms can be trained to predict and compensate for aberrations, reducing reliance on intricate nanoantenna designs. Additionally, computer vision techniques enable the learning and adaptation to ACA-induced degradation patterns, offering a versatile and efficient means to mitigate aberrations in metalens systems. This integration of computer vision enhances the performance and reliability of metalenses and facilitates scalable and adaptable solutions for real-world applications, highlighting the transformative potential of combining computational intelligence with optical design.

\begin{figure}[t]
\begin{center}
\includegraphics[width=1.0\linewidth]{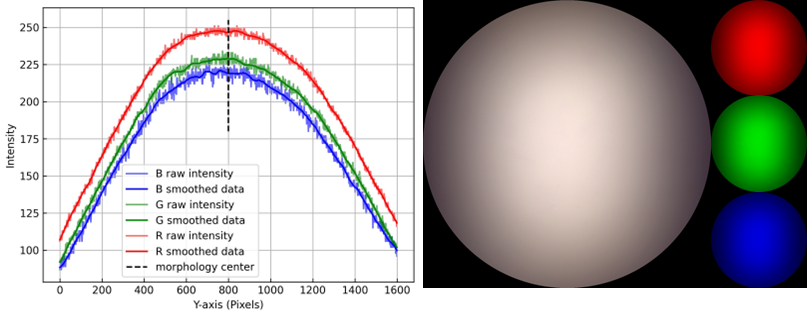} 
\caption{Illustration of the spatial prior of the metalens imaging.}\label{fig:spatial}\vspace{-20pt}
\end{center}
\end{figure}\label{fig:sp_prior}

\begin{figure*}[t]
\begin{center}
\includegraphics[width=0.95\linewidth]{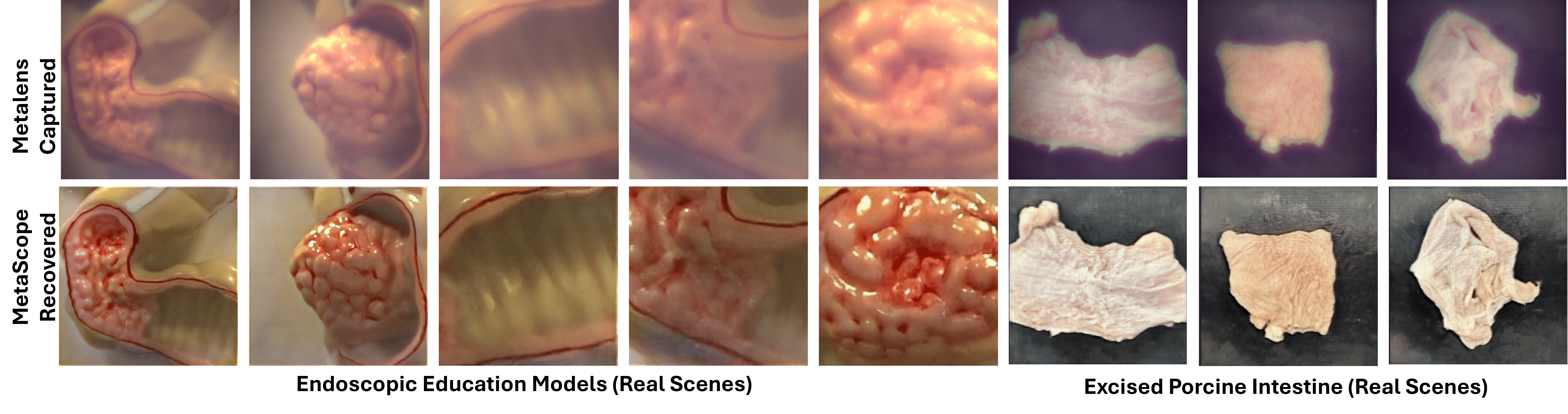} 
\vspace{-10pt}
\caption{Qualitative restoration on the real scenes, including endoscopic education models (Left) and excised porcine intestine (Right). }\label{fig:show_real}
\vspace{-10pt}
\end{center}
\end{figure*}

\begin{table*}[h]
 \centering
\begin{tabular}{l|c|c|c|c|c}
\hline
 & Meta-CVC-Clinic & Meta-CVC-Colon & Meta-Kvasir-Seg& Meta-EndoVis-17 &Meta-EndoVis-18\\ \hline
DRMI & 34.05 & 36.50 & 32.76 & 30.82 & 30.90\\ 
Ours & \textcolor{red}{\textbf{36.47}} & \textcolor{red}{\textbf{37.53}} & \textcolor{red}{\textbf{33.40}} & \textcolor{red}{\textbf{31.40}} & \textcolor{red}{\textbf{31.55}}\\\hline
\end{tabular}
 \caption{Performance comparison on a unified model trained with all five datasets. }
 \label{tab:uni} \vspace{-5pt}
\end{table*}

\begin{table}[t]
 \centering
\begin{tabular}{c|c|c}
\hline
metalens-1 (10 mm) & \multicolumn{2}{c}{metalens-2 (5 mm)} \\\hline
- & Zero-shot & Fine-tuned \\\hline
85.55& 82.82 & \best{86.37} \\\hline
\end{tabular}
\caption{Cross-metalenses (cross-dataset) generalization of using the MetaScope trained on metalens-1 dataset (mIoU). }
\label{tab:cross-data}\vspace{-15pt}
\end{table}

\section{Optical Simulation Experiments}\label{optical}

\subsection{Channel Prior} 
To model the specific color dispersion and derive the channel prior $\mathbf{T}$, we use Fresnel diffraction~\cite{pellat1994fresnel,born2013principles} to simulate the light propagation, as described by the following equation:
\begin{equation}
\begin{split}
 I(x, y, z) &= |E(x, y, z)|^2 \\
 &= \bigg| \iint E_0(u, v) \, P(u, v) \, T(u, v) \, \frac{e^{ikz/\lambda z}}{\lambda z} \\
 &\quad \times \exp\left(\frac{ik}{2z} \left[(x - u)^2 + (y - v)^2\right]\right) \, du \, dv \bigg|^2.
\end{split}
\end{equation}
Here:
\begin{itemize}
\item \( I(x, y, z) \) represents the light intensity at position \((x, y, z)\), which is the energy of electric field $E(x, y, z)$.
\item $E_0(u, v) = A_{0}(u, v) \exp\left(i \phi_0(u, v)\right)$ is the input electric field on the initial plane $(u, v)$. $A_{0}(u, v)$ is the input amplitude, and $\phi_0(u, v)$ is the input phase.
\item $P(u, v) = 
\begin{cases} 
1 & \text{if } \sqrt{u^2 + v^2} \leq R \\
0 & \text{if } \sqrt{u^2 + v^2} > R 
\end{cases}$ is the aperture function. $R$ is the lens radius. An aperture is an optical element that limits the propagation of a light beam, defining the area through which the light passes. The function $P(u, v)$ takes a value of 1 within the aperture area (indicating that light passes) and 0 outside this area (indicating that light does not pass).
\item $T(u, v) = A_{\text{meta}}(u, v, \lambda) \exp\left(i \phi_{meta}(u, v, \lambda)\right)$ is the transformation function of metalens, including the amplitude $A_{meta}(u, v, \lambda)$ and phase $\phi_{meta}(u, v, \lambda)$ modulation provided by the metalens at the position $(u, v)$ for the working wavelength $\lambda$.
\item \( k = \frac{2\pi}{\lambda} \) is the wave vector.
\item \( \lambda \) is the wavelength.
\end{itemize}
With this equation, we could derive the Point Spread Function (PSF) details of all wavelengths through the metalens in 3D space. PSF is the imaging response of an optical system to an ideal point light source, directly governing spatial resolution and color fidelity, which can be regarded as the distortion kernel. The chromatically aberrated image $I_{ca}(x,y)$ can be mathematically expressed as the convolution of the true, clean image with the PSF of the optical system, $I_{ca}(x,y)=I_0(x,y) \ast PSF(x,y)$. For each wavelength, the light intensity \( I(x, y, z) \) will reach the maximum at the focal spot position $(x_0, y_0, f_{\lambda})$. Fig.~\ref{fig:defocus} demonstrates the focal sections along the $x$ axis at the focal plane $(x, y, f)$ for seven wavelengths of different colors, $\sum_{\lambda} I(x, y, z)|_{y=y_0, z=f_{\lambda}}$. Compared to the designed green light at 532 nm, the focusing efficiency of other colors is slightly lower. Tab.~\ref{tab:color_wavelength_efficiency} compares the intensity at the focusing points for each color. This efficiency buffer is the transformation efficiency $\mathbf{T}$ in the proposed Optics-informed Intensity Adjustment (OIA) module. Considering the three-channel properties of images, the red, green, and blue (RGB) ones are encoded into optical embeddings to adjust the channel attention.

\subsection{Spatial Prior} 
To analyze the spatial light distribution pattern $\mathbf{Y}$ of the metalens, a picture of a large bright white object is taken by the metalens. As shown in the right panel of Fig.~\ref{fig:spatial}, we derive a white image whose color is slightly reddish even applying the sensor built-in automatic white balance function with a 1.67:1:2.34 RGB gain. Wavelength-dependent meta-atom responses fundamentally constrain color channel balancing. The sharp exposure boundaries in the white image are defined by the aperture. Threshold-based morphological analysis can yield the morphological center coordinates of the circular area, which is $(x_0, y_0)$. Cutting a line in the white image along $y=y_0$, we can have the spatial RGB light distribution, as shown in the left panel of Fig.~\ref{fig:spatial}. An attenuation exhibits a spatially radial pattern, where the edge regions demonstrate more significant attenuation than the center areas. The radial attenuation pattern stems from two compounding factors: 1) Meta-atom off-axis efficiency decay $\eta_{meta}(\theta)$, where nanostructures exhibit reduced light control capability at oblique angles, and 2) Geometric vignetting governed by the $\cos^4{\theta}$ law. Hence, we directly encode this optical prior at the spatial level to adjust the feature representation.

\begin{figure*}[t]
\begin{center}
\includegraphics[width=0.86\linewidth]{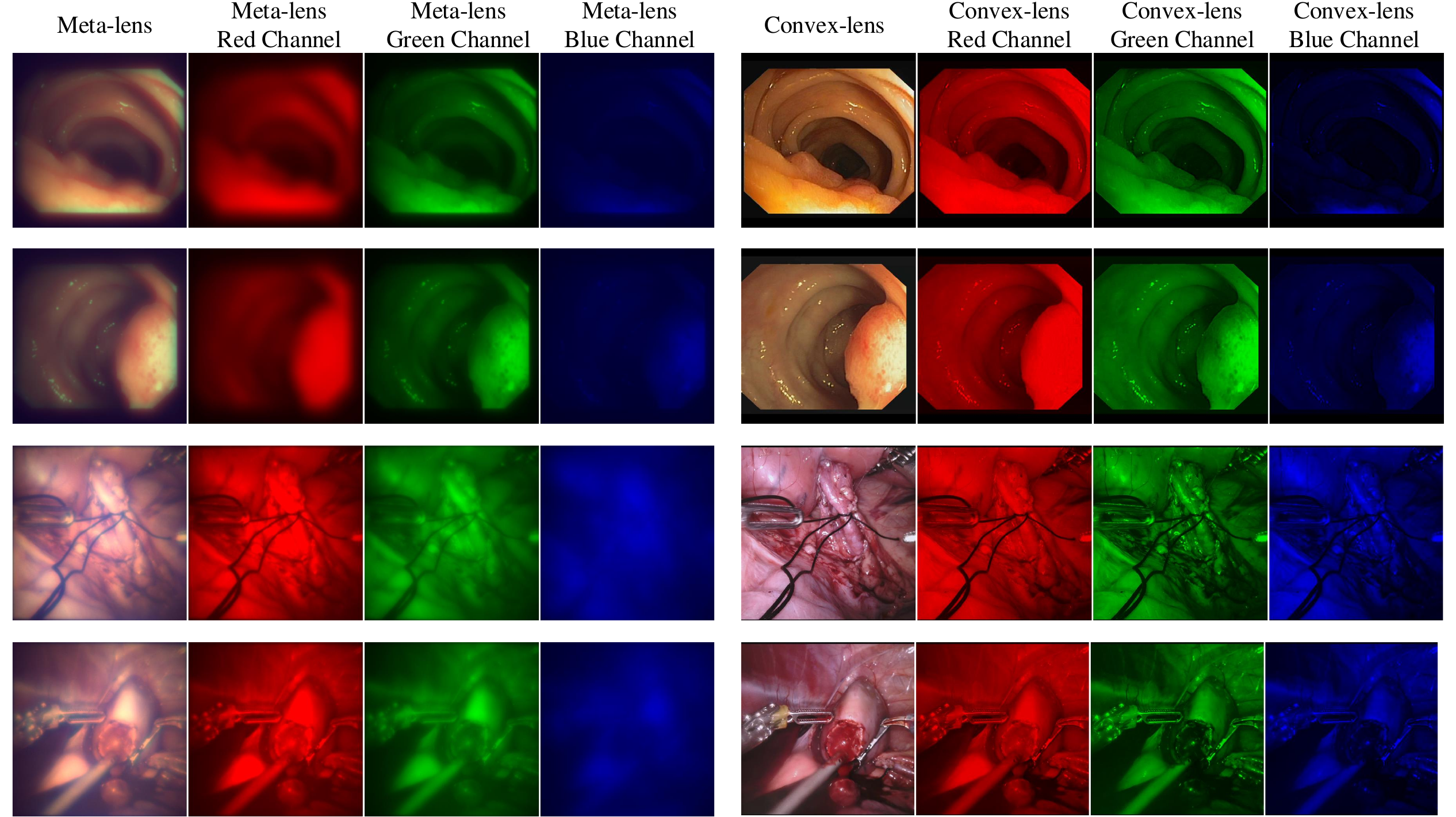} 
\caption{Visualization of the red, green, and blue channels of the samples from our metalens visual datasets. }\label{fig:show_rgb}\vspace{-20pt}
\end{center}
\end{figure*}

\section{More Experimental Verifications} \label{cv}

\subsection{Real Scene Generalization} 

To verify the real-scene generalization, we capture real intestinal scenes using metalens. Due to the unavailability of in vivo experiments, which require additional biological approvals, an endoscopic education model is employed to simulate intestinal scenes. The model is positioned at varying distances (from 1 cm to 7 cm) from the optical system for imaging. As shown in Fig.~\ref{fig:show_real} (Left), MetaScope demonstrates generalization to real scenes and diverse distances, a finding corroborated by~\cite{dong2024achromatic}. Note that capturing real scenes precludes obtaining non-degraded ground truth, limiting model training and quantitative evaluation.

We further conduct biomedical verification by photographing the excised porcine intestine (see Fig.~\ref{fig:show_real} (Right)). Due to the inability to preserve biological tissues for extended periods, this experiment is conducted as a one-time trial. Our MetaScope achieves impressive achromatic correction results, revealing its superiority in clinical generalization. As clinical in vivo verification on humans requires strict ethical certifications, we are actively pursuing the necessary clearance.

\subsection{Scaling Up Training with All Datasets}

To further demonstrate the effectiveness of our algorithmic designs, we train a unified model using all captured data (five datasets) and compare it with the state-of-the-art method in metalens imaging, DRMI~\cite{seo2023deep}. As shown in Tab.~\ref{tab:uni}, our method consistently outperforms DRMI across all datasets, highlighting its superior scalability and generalization capabilities.

\subsection{Generalization to Different Metalenses}
We specifically focus on designing general metalenses that lack achromatic properties, thereby establishing a uniform degradation model. This model is not only relevant to various metalenses but also extends its applicability to diffractive optical elements (DOE) lenses, thereby broadening its scope and significance in optical research.

To validate our approach and dataset, we recaptured the CVC-Clinic dataset using a new metalens with a focal length of 5 mm. We then evaluated the performance of MetaScope on this dataset (Tab.~\ref{tab:cross-data}). Remarkably, without any retraining, MetaScope achieved a satisfactory mIoU of 82.82\%, which is comparable to the mIoU of 85.55\% obtained with the conventional metalens. Furthermore, after fine-tuning, MetaScope demonstrated even superior performance, achieving a mIoU of 86.37\%. These results highlight the significant value of the proposed datasets and the generality of our MetaScope.

\subsection{Model Efficiency}
We further assess the model size and inference speed. MetaScope demonstrates remarkable efficiency, featuring only \textcolor{red}{12M} parameters and achieving an inference speed of \textcolor{red}{23.4 FPS}. This performance significantly outperforms the state-of-the-art Mask2Former, which has \textcolor{red}{44M} parameters and operates at \textcolor{red}{16.9 FPS}. The lightweight architecture of MetaScope enables real-time processing capabilities, making it applicable in endoscopic surgery and diagnostics, where instantaneous imaging and analysis are crucial.

\subsection{Dataset Analysis}

Fig.~\ref{fig:show_rgb} showcases metalens images alongside paired convex-lens images from our meticulously constructed datasets. Compared to convex-lens counterparts, metalens images exhibit noticeable variations in color and blurriness, primarily due to axial chromatic aberration inherent in the metalens design. To explore these in detail, we present separate visualizations for each red, green, and blue channel for both metalens and convex-lens configurations. This comparison reveals significant disparities in intensity degradation and dispersion: while the red and green channels show minimal degradation, the blue channel experiences considerable degradation across all samples. This phenomenon is attributed to the optimization of our metalens for a wavelength of 532 nm with a focal length of 10 mm, which aligns more closely with red and green wavelengths and is less compatible with blue wavelengths (as illustrated in Fig.~\ref{fig:defocus}). These observations have driven the design of our OIA and OCC techniques, which enhance metalens image-based in vivo intelligence. This advancement highlights the superior performance and tailored design of our metalens system in mitigating chromatic aberrations, thereby significantly enhancing reliability for real-world clinical applications.

\subsection{Metalens Imaging Segmentation}

\begin{figure*}[t]
\begin{center}
\includegraphics[width=0.95\linewidth]{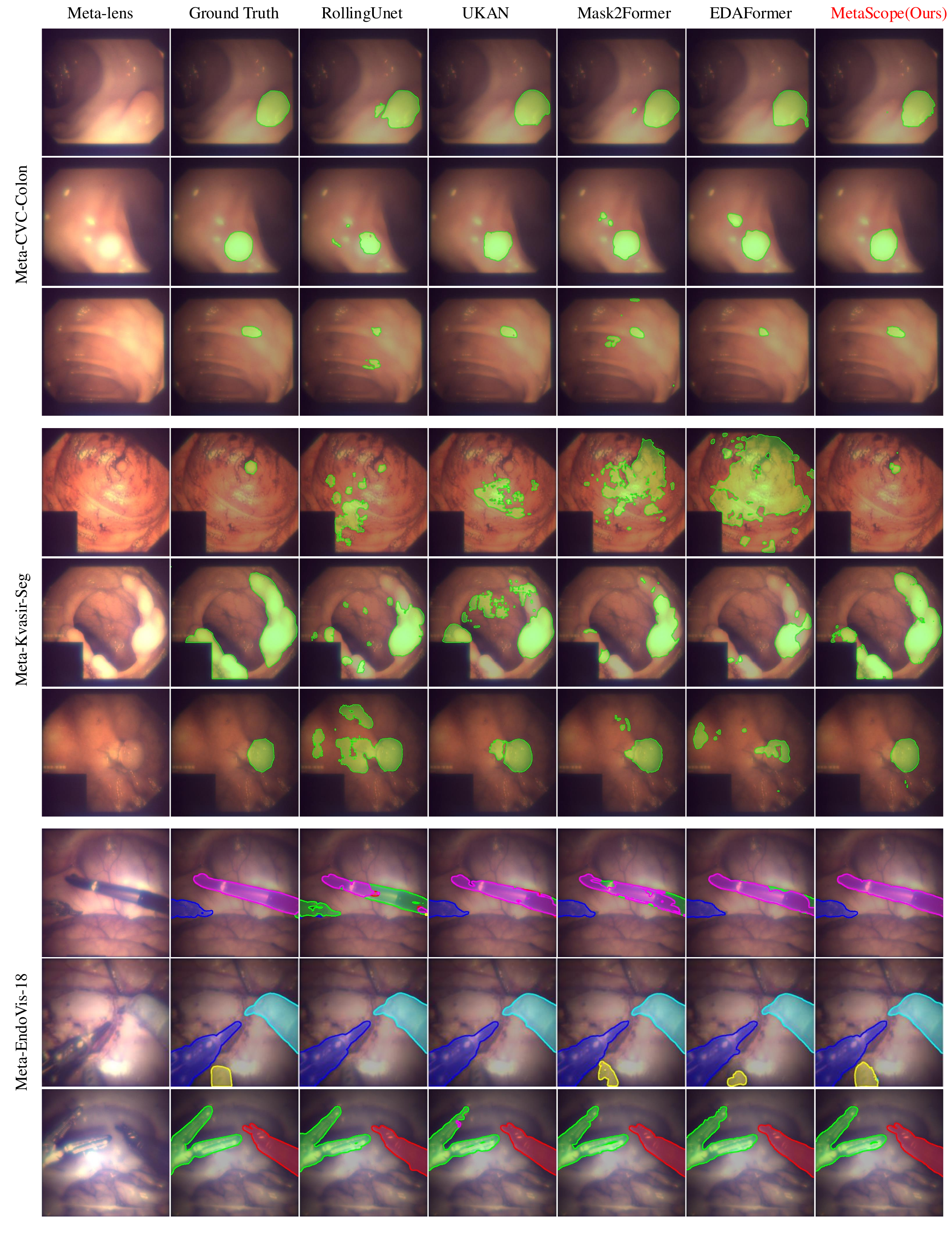} 
\caption{Qualitative comparison of state-of-the-art segmentation methods on Meta-CVC-Colon, Meta-Kvasir-Seg and Meta-EndoVis-18. }\label{fig:show_segmentation}\vspace{-20pt}
\end{center}
\end{figure*}

Fig.~\ref{fig:show_segmentation} presents a qualitative comparison of metalens imaging segmentation performance between our proposed MetaScope and state-of-the-art methods, including Rolling Unet~\cite{liu2024rolling}, U-KAN~\cite{li2024u}, Mask2Former~\cite{cheng2021mask2former}, and EDAFormer~\cite{yu2025embedding}. In abnormality segmentation tasks, as illustrated from row 1 to row 6, MetaScope consistently demonstrates superior accuracy in identifying polyps across various complex scenarios, excelling particularly in detecting small polyps (row 4) and accurately segmenting larger polyp areas (row 5). Furthermore, in the more challenging domain of surgical instrument segmentation, MetaScope continues to outperform current state-of-the-art methods by significantly enhancing the completeness of the ultrasound probe (highlighted in pink in row 7) and accurately delineating large needle drivers (highlighted in red in row 9). Remarkably, MetaScope is also capable of precisely identifying suction instruments (highlighted in yellow in row 8) that are only partially visible within the field of view. These results underscore MetaScope's robust capability to handle complex and partially obscured objects, demonstrating its strong potential and effectiveness for real-world practice.

\subsection{Metalens Imaging Restoration}

\begin{figure*}[t]
\begin{center}
\includegraphics[width=0.85\linewidth]{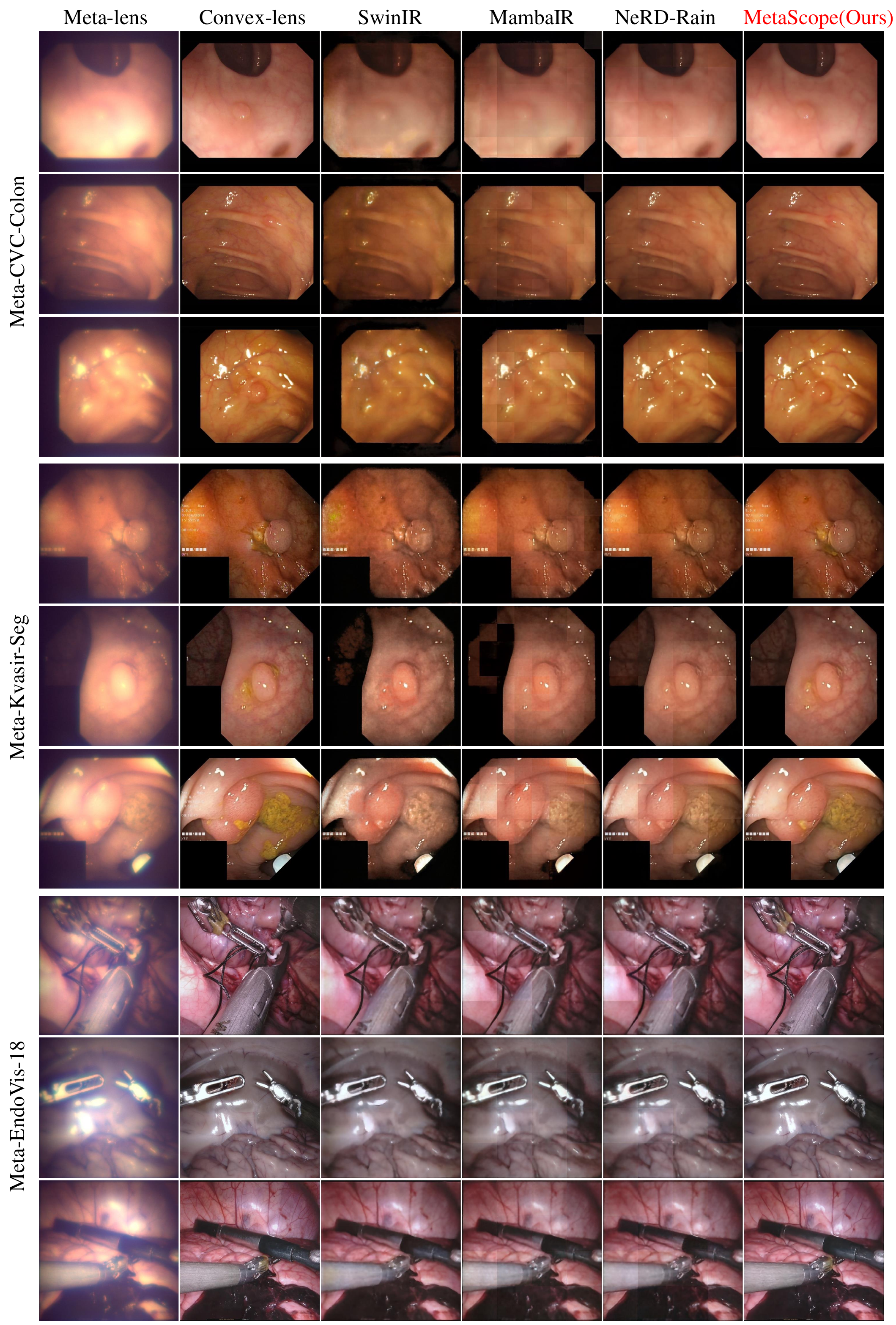} 
\caption{Qualitative comparison with state-of-the-art restoration methods on Meta-CVC-Colon, Meta-Kvasir-Seg and Meta-EndoVis-18. }\label{fig:show_restoration}\vspace{-20pt}
\end{center}
\end{figure*}

Fig.~\ref{fig:show_restoration} provides a visual comparison of restoration quality between our proposed MetaScope and the latest methods, including SWinIR~\cite{liang2021swinir}, MambaIR~\cite{guo2025mambair}, and NeRD-rain~\cite{chen2024bidirectional}. MetaScope significantly outperforms these approaches by accurately restoring the structure and color of polyps (rows 1 to 3), capturing intricate vascular details (row 5), and maintaining the overall scene color fidelity (rows 4 and 6). In more complex surgical scenarios, MetaScope continues to surpass state-of-the-art methods, particularly excelling in rendering accurate colors (rows 7 and 9) and fine texture details (row 8) of surgical instruments. These superior capabilities highlight MetaScope's robust performance in handling diverse and challenging imaging conditions, demonstrating its strong potential and effectiveness for real-world medical applications.

\subsection{Data Visualization}

\begin{figure*}[t]
\begin{center}
\includegraphics[width=0.95\linewidth]{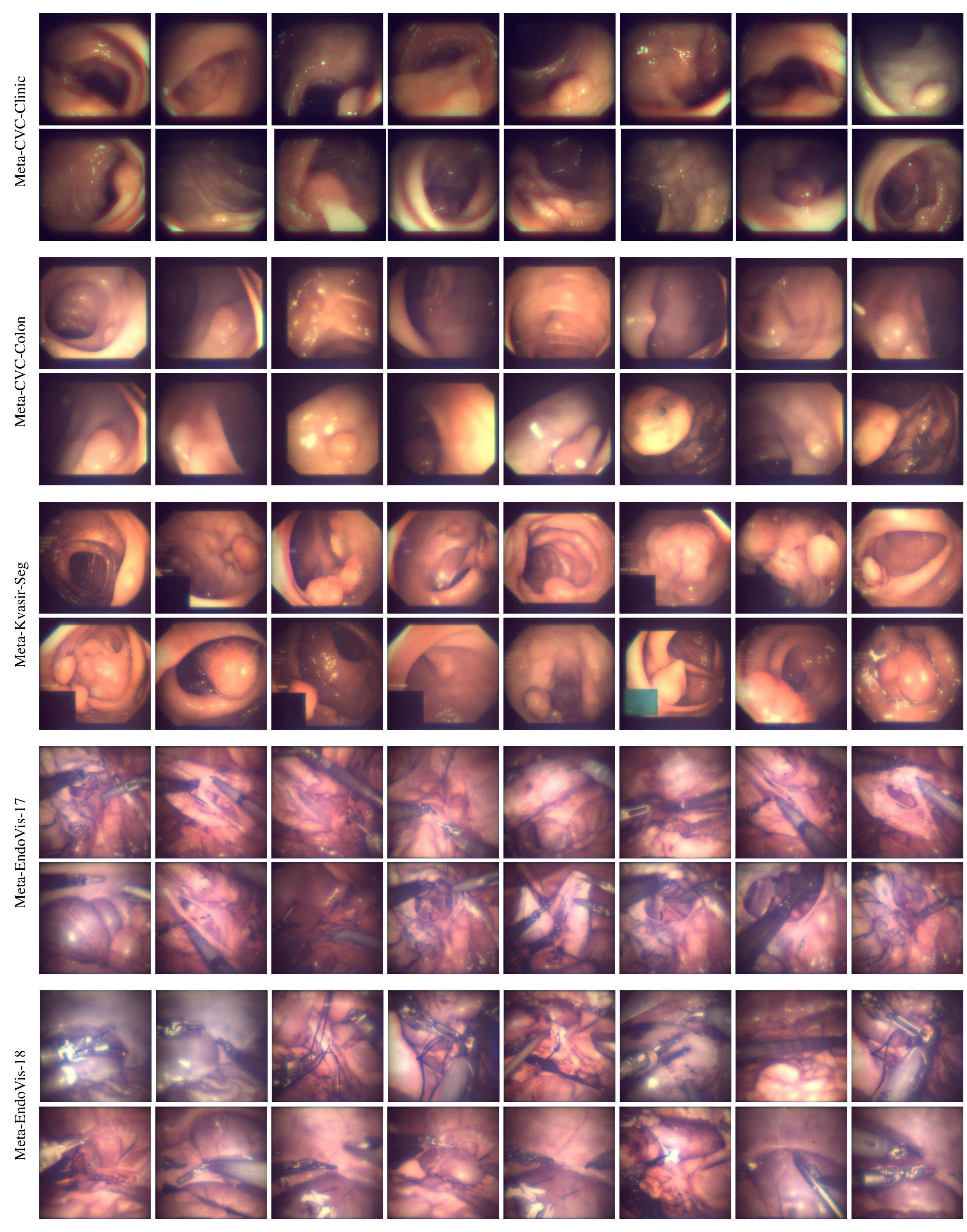} 
\caption{Visualization of samples from five Metalens Imaging datasets.}\label{fig:show_data}\vspace{-20pt}
\end{center}
\end{figure*}

We present sample images from our constructed datasets, including Meta-CVC-Clinic, Meta-CVC-Colon, Meta-Kvasir-Seg, Meta-EndoVis-17, and Meta-EndoVis-18. As illustrated in Fig.~\ref{fig:show_data}, the images across these datasets exhibit similar meta-distortions. Additionally, our datasets offer extensive diversity, encompassing a wide range of in-vivo clinical scenarios that capture various pathological conditions and anatomical variations inherent to endoscopic procedures. This diversity enables comprehensive benchmarking of metalens imaging analysis in clinical in vivo settings, supporting a variety of research objectives, including generalization, transferability, and robustness.

\section{Discussion and Clarifications}\label{discussion}

\subsection{Dataset Information}
The dataset details are as follows: \textbf{(1) CVC-Clinic}~\cite{bernal2015wm} is a publicly available dataset comprising 612 images extracted from 29 colonoscopy videos, each annotated with pixel-wise polyp masks. \textbf{(2) CVC-Colon}~\cite{bernal2012towards} consists of 380 polyp-annotated images sourced from 15 short colonoscopy video sequences. \textbf{(3) Kvasir-Seg}~\cite{jha2020kvasir} is a gastrointestinal polyp segmentation dataset that includes 1,000 images and corresponding segmentation masks. \textbf{(4) EndoVis17}~\cite{allan20192017} consists of 1,800 frames with annotations for various surgical instrument types, enabling the analysis and recognition of instruments in minimally invasive surgeries. \textbf{(5) EndoVis18}~\cite{allan20202018} consists of 2,384 frames and features more complex porcine tissue and dynamic instrument movements with eight classes. Our metalens photoed datasets are named by adding the \textbf{Meta} prefix, such as \emph{Meta-CVC-Clinic}. During dataset construction, we first calibrate the meta-camera to ensure that the photographed and original images are pixel-perfectly aligned~\cite{dong2024achromatic}. Considering the clinical practice with insufficient illumination, we then generate the dataset under artificial light scenarios without sunlight to simulate endoscopic environments.

\subsection{Physically Suitable for Endoscopy}

To achieve accurate diagnosis and surgical operations, high-resolution imaging is essential for providing detailed visualization, such as identifying micro-lesions or vascular structures. Specifications for high-resolution imaging typically require a focal length of \( f = 10 \sim 50 \) mm, a field of view (FOV) of \( 10^\circ \sim 50^\circ \), and an f-number (F/\#) of \( 3.5 \sim 5.6 \).
Our optical system, featuring a focal length of \( f = 10 \) mm, a field of view (FOV) of \( 31^\circ \), and an f-number (F/\#) of 3.8, is fully suitable for endoscopy applications that demand high resolution for observing subtle tissues.

\subsection{Implementation of the KL Divergence }
The proof and implemented version of KL Divergence Loss $L_{\text{KL}}$ (Eq.~7 in the main paper) is detailed as follows. Inspired by variational auto-encoders~\cite{kingma2013auto}, the KL divergence term establishing the proxy offset latent feature $L_{\text{KL}} = \text{KL}\left(q_\phi(\boldsymbol{z} \mid X) | \mathcal{N}(\mathbf{0}, \mathbf{I})\right)$ used in the optics-informed dispersion correction module, is derived as follows:
\begin{equation*}
\begin{aligned}
&\text{KL}\left(q_\phi(\boldsymbol{z} \mid X) | \mathcal{N}(\mathbf{0}, \mathbf{I})\right)\\
= & \int \frac{1}{\sqrt{2 \pi \sigma^2}} e^{-\frac{(x-\mu)^2}{2 \sigma^2}} \log \frac{\frac{1}{\sqrt{2 \pi \sigma^2}} e^{-\frac{(x-\mu)^2}{2 \sigma^2}}}{\frac{1}{\sqrt{2 \pi}} e^{-\frac{x^2}{2}}} \mathrm{~d} x \\
= & \int \frac{1}{\sqrt{2 \pi \sigma^2}} e^{-\frac{(x-\mu)^2}{2 \sigma^2}} \log \frac{1}{\sqrt{\sigma^2}} \times e^{\frac{x^2}{2}-\frac{(x-\mu)^2}{2 \sigma^2}} \mathrm{~d} x \\
= & \int \frac{1}{\sqrt{2 \pi \sigma^2}} e^{-\frac{(x-\mu)^2}{2 \sigma^2}}\left[-\frac{1}{2} \log \sigma^2+\frac{1}{2} x^2-\frac{1}{2} \frac{(x-\mu)^2}{\sigma^2}\right] \mathrm{d} x \\
= & \frac{1}{2} \int \frac{1}{\sqrt{2 \pi \sigma^2}} e^{-\frac{(x-\mu)^2}{2 \sigma^2}}\left[-\log \sigma^2+x^2-\frac{(x-\mu)^2}{\sigma^2}\right] \mathrm{d} x \\
= & \frac{1}{2}\left(-\log \sigma^2+\mathbb{E}\left[x^2\right]-\frac{1}{\sigma^2} \mathbb{E}\left[(x-u)^2\right]\right) \\
= & \frac{1}{2}\left(-\log \sigma^2+\sigma^2+\mu^2-1\right),
\end{aligned}
\end{equation*}
where $\mu$ and $\sigma$ represent the mean and standard deviation of the latent space, respectively. These parameters are learned through two linear layers, $\boldsymbol{\mu}(X)$ and $\log \boldsymbol{\sigma}^2(X)$, as detailed in Section 4.2 of the main paper.

\subsection{Related Work} 
Recent advancements in endoscopic image analysis have significantly enhanced diagnostic capabilities and surgical procedures~\cite{humulti, srivastavprocedure, paruchuri2024leveraging, wu2025self, ma2025shape, liu2025rotation,li2024endora,wuyang2021joint,ali2024assessing,liu2024lgs}. Some works focus on improving representation learning~\cite{humulti, srivastavprocedure,li2022sigma,li2023sigma++,chen2023medical}, removing surgical smoke~\cite{wu2025self}, and integrating complementary modalities~\cite{liu2025rotation} to facilitate precise diagnosis and recognition during inspections and surgeries. Other works target accurate depth estimation~\cite{paruchuri2024leveraging} and pose estimation~\cite{ma2025shape}, and domain shifts~\cite{li2023novel,li2022scan,li2023sigma++} to enable autonomous navigation~\cite{li2025voxdet}. These studies, limited to convex-lens-based systems, hinder the potential for micro-in-vivo intelligence. In contrast, we explore metalens-based perception to advance micro-miniaturized in vivo diagnostics and surgery, fully exploring the inherent optics-driven insights for the methodology design.

\subsection{Ethical Clarification} 
This research complies with all relevant ethical standards and regulations. The data used in this study were sourced from publicly available repositories~\cite{bernal2015wm,bernal2012towards,jha2020kvasir,allan20192017,allan20202018}, ensuring that no identifiable personal information is included. All datasets used adhere to privacy laws and institutional guidelines governing the use of medical information. Additionally, the study does not involve direct interaction with human subjects, eliminating concerns about consent and participant welfare. We confirm that this work does not present ethical issues and aligns with the ethical principles required for medical AI research.


\clearpage

\end{document}